\documentclass[journal]{IEEEtran}

\makeatletter
\def\endthebibliography{%
	\def\@noitemerr{\@latex@warning{Empty `thebibliography' environment}}%
	\endlist
}
\makeatother

\usepackage{amsmath}
\usepackage{amsfonts}
\usepackage{amssymb}

\usepackage{amsthm}
\usepackage{mathtools}
\usepackage{tabularx,booktabs}
\usepackage{graphicx}
\usepackage{subfigure}
\usepackage{enumerate}
\usepackage{float}
\usepackage{url}
\usepackage{verbatim}
\usepackage{cite}
\usepackage{diagbox}
\usepackage{multirow}
\usepackage{makecell}
\usepackage{algorithm}
\usepackage{algorithmic}
\usepackage{bbding}
\usepackage{bm}
\usepackage{cases}
\usepackage[linkcolor=black,citecolor=black,urlcolor=black,colorlinks=true]{hyperref}
\usepackage{tikz}
\usepackage{etoolbox}
\usepackage{graphicx}

\makeatother
\usepackage{amsfonts}

\usepackage{mathtools}
\usepackage{tabularx}
\usepackage{graphicx}
\usepackage{subfigure}
\usepackage{enumerate}
\usepackage{float}
\usepackage{url}
\usepackage{verbatim}
\usepackage[linkcolor=black,citecolor=black,urlcolor=black,colorlinks=true]{hyperref}
\usepackage{hyperref}
\usepackage{array}
\usepackage{multirow}
\usepackage{longtable}
\usepackage{rotating}

\usepackage{color}

\newcommand{\circled}[2][]{\tikz[baseline=(char.base)]
	{\node[shape = circle, draw, inner sep = 1pt]
		(char) {\phantom{\ifblank{#1}{#2}{#1}}};
		\node at (char.center) {\makebox[0pt][c]{#2}};}}
\robustify{\circled}

\renewcommand{\algorithmicrequire}{\textbf{Notation:}}
\renewcommand{\algorithmicensure}{\textbf{Output:}}

\graphicspath{{figures/}}
\IEEEoverridecommandlockouts

\author{Tianyue Wu and Fei Gao\vspace{-0.2cm}\thanks{All authors are with the State Key Laboratory of Industrial Control Technology, Institute of Cyber-Systems and Control, Zhejiang University, Hangzhou, 310027, China. {\tt\small \{tianyueh8erobot, fgaoaa\}@zju.edu.cn}}}

\title{\LARGE \bf Distributed Optimization in Sensor Network for Scalable Multi-Robot Relative State Estimation}

\begin{document}
    \maketitle
    
\begin{abstract}
	Distance measurements demonstrate distinctive scalability when used for relative state estimation in large-scale multi-robot systems. Despite the attractiveness of distance measurements, multi-robot relative state estimation based on distance measurements raises a tricky optimization problem, especially in the context of large-scale systems. Motivated by this, we aim to develop specialized computational techniques that enable robust and efficient estimation when deploying distance measurements at scale. We first reveal the commonality between the estimation problem and the one that finds realization of a sensor network, from which we draw crucial lesson to inspire the proposed methods. However, solving the latter problem in large-scale (still) requires distributed optimization schemes with scalability natures, efficient computational procedures, and fast convergence rates. Towards this goal, we propose a complementary pair of \emph{distributed} computational techniques with the classical \emph{block coordinate descent} (BCD) algorithm as a unified backbone. In the first method, we treat \emph{Burer-Monteiro factorization} as a rank-restricted heuristic for rank-constrained semidefinite programming (SDP), where a specialized BCD-type algorithm that analytically solve each block update subproblem is employed. Although this method enables robust and (extremely) fast recovery of estimates from initial guesses, it inevitably fails as the initialization becomes disorganized. We therefore propose the second method, derived from a \emph{convex} formulation named \emph{anchored \mbox{edge-based} semidefinite programming} (ESDP), to complement it, at the expense of a certain loss of efficiency. This formulation is structurally decomposable so that BCD can be naturally employed, where each subproblem is convex and (again) solved exactly. Since in both methods BCD seeks to solve the subproblem exactly, fewer iterations, as well as the number of communication rounds, are expected. Extensive evaluation on 2D and 3D problems scaling up to tens to hundreds robots show that the proposed methods converge to high-precision results and impressively scale better than the alternative centralized and distributed methods. 
	
\end{abstract}

\section{Introduction}
\label{sec:intro}
\IEEEPARstart{T}hERE is a growing consensus that a large-scale \mbox{multi-robot} system can be more robust, resilient, and efficient in achieving goals that would be difficult to achieve with a single robot \cite{prorok2021beyond, zhou2022swarm}. These multi-robot missions often require individual agents with knowledge of each other's pose and location relative to their own (known as the \emph{relative} \mbox{\emph{state} \cite{cossette2021relative,cossette2022optimal,shalaby2021relative,nguyen2023relative,xu2020decentralized,ziegler2021distributed,xu2022omni}}) to collaborate on the downstream task.

\emph{Inter-robot} measurement is powerful to generate \emph{a priori} for consistent relative state estimation. However, once the system is scaled up, the physical setup generating these measurements has to be judiciously chosen. For example,  (direct) inter-robot visual  measurements \cite{xu2020decentralized,xu2022omni} suffer from the problem of occlusion and demanding ID identification of individual agents, characteristics that become irresistible in dense, large-scale systems. Environmental (indirect) relative measurements (e.g., inter-robot loop closures \cite{giamou2018talk}) cause perceptual aliasing that confuses robots with similar-appearance scenarios, which generally creates a significant computational burden in large-scale systems for robust data association \cite{tian2022kimera,wu2023decoupled}. Fortunately, the abovementioned problems can be greatly alleviated by some modern distance sensors (e.g., ultra-wideband (UWB)) \cite{nguyen2023relative}. They are generally lightweight and inexpensive while enabling long-range (tens to hundreds of meters), large-capacity (number from tens to hundreds) and high-precision (centimeter-level) measurements. 

Despite the scalability benefiting from distance sensors, the distance-based relative state estimation araising from large-scale systems implies an intractable optimization problem. The high-dimensional, non-convex search space as well as the presence of Euclidean distances make it tricky to solve the problem. To ease the embarrassment, our key insight is to explore the relationship between relative states among robots and the sensors' point coordinates. By doing so, we are able to (i) search for solutions in a linear Euclidean space with computationally tractable constraints, and (ii) relate the state estimation problem to the rich literature of the one that finds \emph{point realizations/conformations} such as sensor network \mbox{localization\cite{so2007theory,biswas2006semidefinite,biswas2006distributed,tseng2007second,wang2008further,pong2011robust,nishijima2022block}}, molecular \mbox{conformation\cite{crippen1988distance,yoon2000mathematical,biswas2008distributed,leung2010sdp}, etc.} By extending the technical tools and theoretical or empirical evidence provided by these works, we are able to present a reasonable view of designing the practical methods that ease the computational issues and retain the scalability benefits of distance measurement. 

While the centralized approach is a straightforward choice for relative state estimation, its runtime typically rises at least at a linear rate with the dimensionality of the problem \cite{nesterov2012efficiency} (which is proportional to the scale of the system), and thus not scalable. In contrast, \emph{distributed optimization} thus seems a sure path to scalability, where ideally, individual agents work on their local problems and communicate only with their \mbox{pre-defined} neighbors. However, distributed architectures typically require robots to communicate iteratively, where, despite the overload per communicating in distributed optimization being small compared to its centralized counterpart, more communication rounds\cite{arjevani2015communication} tend to result in less robust algorithmic practices. In this paper, we adopt the classical
block coordinate descent (BCD) algorithm [36], which is one
of the natural choices for distributed and scalable optimization
techniques, as a unified backbone for our proposed methods.
In the proposed methods, the BCD algorithm solves each
block update subproblem exactly as a (most) direct way to
reduce the number of iterations thereby alleviating potential
communication fragility. Each block update subproblem in
the proposed methods is well-structured such that allows
for efficient implementations of the usually unavailable or
expensive exact solution step. In particular, we propose (i) an unconstrained nonlinear programming model that derived from \emph{Burer-Monteiro factorization} \cite{burer2003nonlinear}, which allows a extremely efficient solution procedure using a BCD algorithm with close-formed updates, and (ii) a convex formulation based on \emph{edge-based SDP} (ESDP) \cite{wang2008further} that gains robustness to initializations and is structurally decomposable such that allows efficient solution using BCD.

In summary, we make the following contributions: 
\begin{itemize}
	\item [1)] 
	We present a novel perspective for the distance-based relative state estimation problem by bridging it with point realization problems.
	\item [2)] 
	We propose two distributed computational techniques that are theoretically unified (the proofs of Theorem 1 and Theorem 2) and practically complement each other (Figure 2) with their different computational cost and requirement for initial guesses. The first method enables fast  recovery of relative state estimates from chaotic initializations in a large-scale system, while the second method futher boosts the robustness to initializations, at the expense of generally taking more time.
	\item [3)] 
	Evaluations showcase that the proposed methods almost comprehensively (in terms of accuracy, efficiency, and robustness) dominate other centralized or distributed alternatives that demonstrate unacceptable performance in large-scale. In comparison, the proposed methods perform considerably well, making the untrivial distance-based relative state estimation in large-scale possible.
\end{itemize}

\section{Related Work}
\subsection{Relative State Estimation}
\label{sec:related}
Fusion of intra-robot (e.g., odometry) and inter-robot measurements is a popular paradigm for relative state estimation. Xu et al. \cite{xu2020decentralized,xu2022omni} fuses vision-inertial odometry (VIO), visual-detection, UWB measurements and even environmental features to achieve accurate relative state estimation for small drone swarms. However, these approaches, in addition to the problems mentioned in Section \ref{sec:intro}, require transmission of a large-volume of measurements, causing a heavy communication overload. Moreover, these methods require complex and expensive hardware designs that are not practically scalable. Nguyen et al. \cite{nguyen2023relative} propose QCQP and SDP formulations for the relative transformation estimation problem in 4-DoF (assuming known roll and pitch angles measurements by IMUs) of VIO-UWB fusion. Similar ideas of using semidefinite programming for such problems can be also found in \cite{jiang20193,li20223}. However, the setup in \cite{nguyen2023relative} leads the problem of observability and the proposed approaches are only applicable to a single pair of robots. Ziegler et al. \cite{ziegler2021distributed} are one of the few examples that employ distributed optimization techniques (in their work, ADMM) for estimating relative states using VIO-UWB fusion.  However, the formulation in their work is too sensitive to the initial guesses and prior estimates of odometries for robust estimation. Cornejo et al. \cite{cornejo2015distributed} study the problem of relative state estimation using odometry and distance measurements in the 2D case, where they propose distributed algorithms and analyze the degenerate configurations in this setup. However, due to the simplification of the problem, the method proposed in \cite{cornejo2015distributed} does not guarantee consistency among the estimates (i.e., the estimation of a robot's state may be different for different robots), and thus is not as robust to noises as the consistent ones.

To address the general problems araising from data transmission and observability in the above intra-inter measurement fusion paradigm, some works \cite{cossette2021relative,shalaby2021relative,cossette2022optimal} propose to equip each robot with multiple distance sensors to achieve an odometry-free relative localization approach. Although the setup considered in this paper is similar to these works, they haven't explicitly address the computational issues araising in the underlying optrtimization problem, especially in the context of large-scale; moreover, the techniques introduced in this paper can be directly applied to the setup in those works and thus contribute in a parallel (complementary) way.
\subsection{Semidefinite Programming for Robot Estimation}
Recently, semidefinite relaxations (SDR) have been successfully applied to the Pose Graph Optimization (PGO) problem in robotics, where SE-Sync \cite{rosen2019se} and its distributed variants \cite{tian2021distributed} use the \emph{Riemannian Staircase} approach \cite{boumal2015riemannian,boumal2016non} that allows to solve SDPs efficiently using nonlinear programming (NLP) algorithms. This procedure can be carried out by solving a series of NLP problems through a "optimize-certify" manner until the solution of the relaxed SDP is recovered, where the termination condition is guaranteed to be reached. A feature of the PGO problem is that, under mild preconditions, the solution of the relaxation problem is fortunately equal to the one of the original PGO (i.e., the relaxtion is tight) \cite{bandeira2017tightness}. The problem considered in this paper, however, is  substantially different from the PGO problem in (at least) the following aspects: (i) the objective function of the relaxation counterpart for our problem is nonlinear, unlike the ones in \cite{rosen2019se,tian2021distributed}, (ii) the relaxation in this paper is not tight at all with high-rank solutions in its solution set, and (iii) the search space does not enjoy the distinctive geometry as in \cite{rosen2019se,tian2021distributed,boumal2015riemannian}. We therefore consider different algorithmic designs instead of the Riemannian Staircase: our proposed NLP model (Problem 4) treats Burer-Monteiro (BM) factorization, also used in the Riemannian Staircase, as a rank-restricted heuristic for fast local searching \cite{burer2003nonlinear}; the proposed convex formulation (Problem~5) is no longer a Shor’s relaxation \cite{rosen2019se} of the original problem,which is in contrast a further relaxation \cite{wang2008further} of the direct rank relaxation for the original problem, which facilitates the use of distributed optimization algorithms. We note that while the previous work \cite{tian2021distributed} also employ a BCD-type distributed solver that employ approximate update for each subproblem derived from BM factorization \cite{erdogdu2022convergence}, the BCD algorithms in this paper solve each subproblem exactly.

\section{From Relative State Estimation to Realization of Sensor Network}
\vspace{-0.1cm}
\label{Sec2}
In this section, we specify the setup of distance-based relative state estimation and transform it into a problem of finding realizations of a sensor network, which can be formulated as a rank-constrained SDP. We then analyze the dilemma in addressing this problem on the perspective from the literature related to point realization problems. 
\vspace{-0.16cm}
\subsection{Problem Setup and Preliminaries}
\vspace{-0.08cm}
To ensure that the relative states are observable in most cases, we consider a multi-robot system with $n$ members equipped with \emph{two} distance sensors\footnote{Note that the results in the latter sections can be extended to heterogeneous configurations, where the number of distance sensors on different robots varies, requiring only additional considerations of observability \cite{shalaby2021relative}.} and being able to obtain accurate estimates of their pitch \mbox{angle $\theta$} and roll \mbox{angle $\phi$}, e.g., from IMUs. The latter assumption is common in related literature, e.g., \cite{shalaby2021relative,nguyen2023relative,ziegler2021distributed}. We specify a \emph{common reference system}, in which the robots' states and the coordinates of the sensors are expressed. Since we are considering the problem of relative state estimation, the common reference system can be set arbitrarily when $d=2$, while its z-axis should be aligned with the opposite direction of gravity when $d=3$, where $d$ is the dimension of the problem. In the following, we show the results for case $d=3$ by default.

We model the measurements topology underlying the relative state estimation using an undirected graph $\mathcal{G} =\left( \mathcal{V} ,\mathcal{E} \right)$ in which each node $i\in \mathcal{V}$ represents one robot in the system and the edges $(i,j) \in \mathcal{E}$ indicate that the measurements between \mbox{robots $i$, $j$}. A binary set $\mathcal{B} \coloneqq \left\{ 0,1 \right\} $ is introduced to distinguish between the two distance sensors on a robot. Let \mbox{$\boldsymbol{p}\coloneqq\left[ p_{1}^{0},p_{1}^{1},p_{2}^{0},p_{2}^{1},...,p_{n}^{0},p_{n}^{1} \right]$ $\in \mathbb{R} ^{d\times 2n}$}
denote a realization of points in $\mathbb{R} ^d$. Each component  \mbox{$p_{i}^{u}\coloneqq\left[ x_{i}^{u},y_{i}^{u},z_{i}^{u} \right]^T\in~\mathbb{R} ^3$} for \mbox{$i=1,2,...,n$} and $u\in \mathcal{B}$, is the coordinate of sensor $(i,u)$. Letting $\widetilde{\left( \cdot \right) }$ denote measurements and $\underline{\left( \cdot \right) }$ denote the (latent) ground-truth, we have the following measurement model for the distance between sensor $(i,u)$ and $(j,v)$:
\vspace{-0.1cm}
\begin{equation}
	\begin{split}
		\label{eq:measurement model0}
		\tilde{d}_{ij}^{uv}=\underline{d}_{ij}^{uv}+\epsilon=\left\| \underline{p}_{i}^{u}-\underline{p}_{j}^{v} \right\| +\epsilon,  \hspace{0.2cm}\epsilon \sim \mathcal{N} \left( 0,\sigma^2 \right), \\ 
		\forall \left( i,j \right) \in \mathcal{E},\ u,v\in \mathcal{B},
	\end{split}	 
	\vspace{-0.3cm}
\end{equation} 
where $\epsilon$ is the measurement noise and $\sigma$ is the noise level. For optimization reasons, we consider the following quadratic measurement model~\cite{trawny2010global}:
\vspace{-0.15cm}
$$
\label{eq:measurement model^2}
\tilde{q}_{ij}^{uv}\coloneqq\left( \tilde{d}_{ij}^{uv} \right) ^2-\sigma ^2\simeq \left( \underline{d}_{ij}^{uv} \right) ^2+\epsilon _{ij}^{uv}, \epsilon _{ij}^{uv}\sim \mathcal{N} \left( 0,(\sigma_{ij}^{uv})^2\right),    
$$
\vspace{-0.4cm}
\begin{equation}
	\sigma_{ij}^{uv}=\sqrt{\left(2\sigma \tilde{d}_{ij}^{uv} \right) ^2+2\sigma ^4}, \ \ 
	\forall \left( i,j \right) \in \mathcal{E},\ u,v\in \mathcal{B}.
\end{equation}

Letting $\underline{R}_i\in \mathrm{SO}\left( d \right) $, $\underline{t}_i\in \mathbb{R} ^d$
represent the rotation and translation components of the robot $i$'s state in the common reference system and $\bar{\nu}_{i}^{u}$ the coordinate of \mbox{sensor $u$} on the robot $i$ in its body reference system, we immediately have
\begin{equation}
	\label{eq:p=Rv+t}
	\underline{p}_{i}^{u} = \underline{R}_i\bar{\nu}_{i}^{u}+\underline{t}_i.
\end{equation}
Given a set of noisy measurements $\tilde{q}_{ij}^{uv}$, it is straightforward to derive the \mbox{maximum-likelihood} estimation (MLE) model for 3D distance-based relative state estimation: \vspace{0.1cm} \\ 
\textbf{Problem 1} (MLE model for Euclidean distance-based relative state estimation).
\vspace{-0.2cm}
\begin{equation}
	\underset{\left\{ (R_i, t_i) \right\}}{\min}\hspace{-0.2cm}\hspace{-0.1cm}\sum_{\footnotesize \begin{array}{c}
			\left( i,j \right) \in \mathcal{E}\\
			u,v\in \mathcal{B}\\
	\end{array}}{\hspace{-0.2cm}\frac{1}{ (\sigma _{ij}^{uv})^2 }\left( \left\| R_i\bar{\nu}_{i}^{u}+t_i-R_j\bar{\nu}_{j}^{v}-t_j \right\|^2 \hspace{-0.15cm}-\tilde{q}_{ij}^{uv} \right) ^2},
	\vspace{-0.3cm}
\end{equation}
\ \ \  s.t. 
\vspace{-0.5cm}
\begin{equation}
	\label{eq:R constraint}
	\begin{split}
		R_i=R_{i|z}\tilde{R}_{i|y}\tilde{R}_{i|x} 
		=\left[ \begin{matrix}
			\cos \psi _i&		-\sin \psi _i&		0\\
			\sin \psi _i&		\cos \psi _i&		0\\
			0&		0&		1\\
		\end{matrix} \right] * 
		\\ \left[ \begin{matrix}
			\cos \tilde{\phi}_i&		0&		\sin \tilde{\phi}_i\\
			0&		1&		0\\
			-\sin \tilde{\phi}_i&		0&		\cos \tilde{\phi}_i\\
		\end{matrix} \right] *\left[ \begin{matrix}
			1&		0&		0\\
			0&		\cos \tilde{\theta}_i&		-\sin \tilde{\theta}_i\\
			0&		\sin \tilde{\theta}_i&		\cos \tilde{\theta}_i\\
		\end{matrix} \right],
	\end{split}
	\vspace{-0.2cm}
\end{equation}
where ${\tilde{\phi}_i}$ and ${\tilde{\theta}_i}$ are the estimates of pitch and roll angles, while $\psi_i$ is the yaw angle to be estimated. 
\vspace{-0.35cm}
\subsection{Transformation of Problem 1}
\vspace{-0.1cm}
Next we present the results that utilizes point coordinates of the sensors as decision variables instead of the robots' poses. First, we will show that the sensor coordinates $\{(\underline{p}_{i}^{u},\underline{p}_{i}^{v})\}$ are in \mbox{one-to-one} correspondence with the states  $\{(\underline{R}_{i},\underline{t}_{i})\}$ to showcase the equivalence of the substitution. 

From (\ref{eq:p=Rv+t}), we have \vspace{-0.2cm}
\begin{equation}
	\label{eq:dp=Rv}
	\begin{split}
		p_{i}^{u}-p_{i}^{v}=R_i\left( \bar{\nu}_{i}^{u}-\bar{\nu}_{i}^{v} \right) \ \ \ \ \ \ \ \ \ \ \ \ 
		\\ =\left[ \begin{matrix}
			\cos \psi _i&		-\sin \psi _i&		0\\
			\sin \psi _i&		\cos \psi _i&		0\\
			0&		0&		1\\
		\end{matrix} \right] *\left( \tilde{R}_{i|y}\tilde{R}_{i|x}\left( \bar{\nu}_{i}^{u}-\bar{\nu}_{i}^{v} \right) \right). 
	\end{split}
	\vspace{-1.2cm}
\end{equation}
Introducing $
\left[ \bar{\nu}_{i|x};\bar{\nu}_{i|y};\bar{\nu}_{i|z} \right]\coloneqq \tilde{R}_{i|y}\tilde{R}_{i|x}\left( \bar{\nu}_{i}^{u}-\bar{\nu}_{i}^{v} \right)$ for the sake of concise writing, a straightforward computation shows the following relations from the first two rows of (\ref{eq:dp=Rv}):
\vspace{-0.25cm}
\begin{equation}
	\label{eq:sinpsi}
	\sin \psi _i={{\left( \bar{\nu}_{i|x}\left( y_{i}^{u}-y_{i}^{v} \right) -\bar{\nu}_{i|y}\left( x_{i}^{u}-x_{i}^{v} \right) \right)}\Bigg/{\left( {\bar{\nu}_{i|x}}^2+{\bar{\nu}_{i|y}}^2 \right)}},
\end{equation}
\vspace{-0.5cm}
\begin{equation}
	\label{eq:cospsi}
	\cos \psi _i={{\left( \bar{\nu}_{i|x}\left( x_{i}^{u}-x_{i}^{v} \right) +\bar{\nu}_{i|y}\left( y_{i}^{u}-y_{i}^{v} \right) \right)}\Bigg/{\left( {\bar{\nu}_{i|x}}^2+{\bar{\nu}_{i|y}}^2 \right)}},  
\end{equation}
So far a one-to-one correspondence between $\{(\underline{p}_{i}^{u},\underline{p}_{i}^{v})\}$ and $\{(\underline{R}_{i},\underline{t}_{i})\}$ is found.

Now it is ready to establish the MLE model based on the coordinates of the sensor network. Since $\psi$ should be rendered redundant in this model, we integrate (\ref{eq:sinpsi}), (\ref{eq:cospsi}) using $\sin ^2\psi _i+\cos ^2\psi _i=1$, to obtain the constraint
\vspace{-0.2cm}
\begin{equation}\nonumber
	\left\| p_{i}^{u}-p_{i}^{v} \right\| = \left\| \bar{\nu}_{i}^{u}-\bar{\nu}_{i}^{v} \right\|,
	\vspace{-0.2cm}
\end{equation}
and the last row of (\ref{eq:dp=Rv}) implies that\vspace{-0.2cm}
\begin{equation}\nonumber
	\begin{split}		
		z_{i}^{u}-z_{i}^{v}=\left[ \begin{matrix}
			-\sin \tilde{\phi}_i,\ \cos \tilde{\phi}_i\sin \tilde{\theta}_i,\ \cos \tilde{\phi}_i\cos \tilde{\theta}_i\\
		\end{matrix} \right] \left( \bar{\nu}_{i}^{u}-\bar{\nu}_{j}^{v} \right)\hspace{-0.05cm}. 
	\end{split}
	\vspace{-1.8cm}
\end{equation} 
Summarizing the above equations (along with (\ref{eq:p=Rv+t})), we can obtain the following equivalent problem of Problem 1:
\vspace{0.05cm} \\ 
\textbf{Problem 2} (MLE model for robotic sensor network localization).
\vspace{-0.3cm}
\begin{equation}
	\label{eq:problem 1}
	\underset{\boldsymbol{p}}{\min}\sum_{\footnotesize \begin{array}{c}
			\left( i,j \right) \in \mathcal{E}\\
			u,v\in \mathcal{B}\\
	\end{array}}{\frac{1}{ (\sigma _{ij}^{uv})^2 }\left( \left\| p_{i}^{u}-p_{j}^{v} \right\|^2 -\tilde{q}_{ij}^{uv} \right) ^2},
	\vspace{-0.4cm} 
\end{equation}
\vspace{-0.2cm} s.t. \vspace{-0.1cm}
\begin{equation}
	\label{eq:norm constraint}
	\left\| p_{i}^{u}-p_{i}^{v} \right\| =\left\| \bar{\nu}_{i}^{u}-\bar{\nu}_{i}^{v} \right\|, \ \forall i\in \mathcal{V} , u,v\in \mathcal{B} , u< v,
	\vspace{-0.05cm}
\end{equation}
\begin{equation}
	\label{eq:z constriant}
	\begin{split}		
		z_{i}^{u}-z_{i}^{v}=\left[ \begin{matrix}
			-\sin \tilde{\phi}_i,\	\cos \tilde{\phi}_i\sin \tilde{\theta}_i, \		\cos \tilde{\phi}_i\cos \tilde{\theta}_i\\
		\end{matrix} \right] \left( \bar{\nu}_{i}^{u}-\bar{\nu}_{j}^{v} \right)\hspace{-0.05cm}, 
	\end{split}
\end{equation}
\vspace{-1.6cm}
\begin{center}
	$$\forall u,v\in \mathcal{B} , u< v.$$
\end{center}
\vspace{-0.2cm}

It is worth noting that Problem 2 shares the same objective function as the (anchor-free) noisy \emph{sensor network localization} problem \cite{biswas2006semidefinite} with additional constraints~(\ref{eq:norm constraint}), (\ref{eq:z constriant}).
\subsection{Problem~2 as a Rank-constrained SDP}
Problem 2 allows a standard rank-constrained semidefinite programming (SDP) formulation with a corresponding relaxed SDP problem    \cite{so2007theory, biswas2006semidefinite}. To reveal this, variable $X$ is critically introduced as follows:
\begin{equation}
	X\coloneqq \boldsymbol{p}^T\boldsymbol{p}.
\end{equation}
The above substitution is equivalent to 
$$
X\succeq \boldsymbol{p}^T\boldsymbol{p}, \ \mathrm{rank}\left( X \right) \leq d,  \vspace{0.cm}
$$
and thus can be rewritten as \cite{boyd1994linear}
\begin{equation}
	\label{eq:Z>0}
	Z\coloneqq \left[ \begin{matrix}
		X&		\boldsymbol{p}^T\\
		\boldsymbol{p}&		I_d\\
	\end{matrix} \right] \succeq 0,\ \mathrm{rank}\left( Z \right) \leq d.
\end{equation}
It is readily shown that  \vspace{-0.1cm}
\begin{equation}
	\begin{split}
\left\| p_{i}^{u}-p_{j}^{v} \right\| ^2=Z_{2i-1+u,2i-1+u}-Z_{2i-1+u,2j-1+v} \\ -Z_{2j-1+v,2i-1+u}+Z_{2j-1+v,2j-1+v}=\mathcal{A} _{ij}^{uv}\cdot Z,  
	\end{split}
\end{equation}
\begin{equation}
	\begin{split}
		z_{i}^{u}-z_{i}^{v}=X_{2n+3,2i-1+u}\hspace{-0.05cm}-\hspace{-0.05cm}X_{2n+3,2i-1+v}=\mathcal{A} _{i|z}\hspace{-0.05cm}\cdot\hspace{-0.05cm} Z,
	\end{split}
\end{equation}
where operator $\mathcal{A} _{ij}^{uv}$ and $\mathcal{A} _{i|z}$ are defined to capture the above linear mapping. From above, it is directly to declare that Problem 2 can be formulated as a SDP with a rank constraint appearing in~(\ref{eq:Z>0}), as follows:\vspace{0.1cm} \\ 
\textbf{Problem 3} (Rank-constrained SDP for Problem 2).
\begin{equation}
	\underset{Z}{\min}\sum_{\footnotesize\begin{array}{c}
			\left( i,j \right) \in \mathcal{E}\\
			u,v\in \mathcal{B}\\
	\end{array}}\frac{1}{ (\sigma _{ij}^{uv})^2 }{\left( \mathcal{A} _{ij}^{uv}\cdot Z-\tilde{q}_{ij}^{uv} \right) ^2},
\end{equation}
\ \  \hspace{0.1cm} s.t. 
$$
Z\coloneqq \left[ \begin{matrix}
	X&		\boldsymbol{p}^T\\
	\boldsymbol{p}&		I_d\\
\end{matrix} \right] \succeq 0,\ \mathrm{rank}\left( Z \right) \leq d.
$$
\begin{equation}
	\label{eq:norm constraint A}
	\mathcal{A} _{ii}^{uv}\cdot Z=\left\| \bar{\nu}_{i}^{u}-\bar{\nu}_{i}^{v} \right\|^2, \ \forall i\in \mathcal{V} , u,v\in \mathcal{Z} , u< v,
\end{equation}
\begin{equation}
	\label{eq:z constraint A}
	\begin{split}
		\mathcal{A} _{i|z}\cdot Z=\left[ \begin{matrix}
			-\sin \tilde{\phi}_i,\	\cos \tilde{\phi}_i\sin \tilde{\theta}_i, \		\cos \tilde{\phi}_i\cos \tilde{\theta}_i\\
		\end{matrix} \right] \left( \bar{\nu}_{i}^{u}-\bar{\nu}_{j}^{v} \right), 
	\end{split}
\end{equation}
\vspace{-1.7cm}
\begin{center}
	$$\forall u,v\in \mathcal{B} , u< v.$$
\end{center}
\vspace{-0.1cm} 

This formulation clearly captures the form of the problem: minimizing a quadratic objective function, being subject to semidefinite and affine constraints, and having a feasible set that are nonconvex due to the rank constraint. This problem is difficult to solve directly for a KKT point\cite{sun2017rank}, not to mention the global minimizer.

\subsection{The Delimma in Addressing Problem 3 in Practice}
A classic approach in related literature \cite{so2007theory,biswas2006semidefinite,biswas2006distributed,biswas2008distributed,leung2010sdp,rosen2019se,tian2021distributed} to general rank-constrained SDPs is to drop the rank constraint to obtain convex surrogates of them. If the objective function is nonlinear, as in Problem 3, in the presence of measurement noise we usually obtain a high-rank (often much higher than $d$) solution by solving the relaxation problem, meaning that the relaxation is not tight. As a SDP, this problem can in principle be solved in polynomial time using standard convex programming techniques, e.g., interior point methods; however, the high computational cost and the centralized nature of generic algorithms can limit their scalability in practice.

In addition to the computational issues, another difficulty is to rounding an informed estimate of the original problem from the solution of relaxation problem. One possible way is to directly extract $\boldsymbol{p}$ from $Z$ from the \mbox{(2, 1)} block of the $Z$ \mbox{in~(\ref{eq:Z>0})}. However such an approach has been shown in previous work not to work well for problems without sufficient anchors~\cite{biswas2006semidefinite,biswas2008distributed,leung2010sdp}. Another way is to perform an decomposition on $X$ such that to obtain an approximate low-rank \mbox{solution \cite{yoon2000mathematical,biswas2008distributed,leung2010sdp}}. In particular, $X$ can be eigenvalue decomposed such that $X=Q\varLambda Q^{-1}$ where $Q$ is the eigenvector matrix of $X$ and $\varLambda$ is the eigenvalue (diagonal) matrix. An approximate rank-$d$ factorization $\hat{\boldsymbol{p}}$ of $X$ can be obtained as follows: 
\begin{equation}
X\simeq \hat{\boldsymbol{p}}^T\hat{\boldsymbol{p}},
\end{equation}
with
$$
\hat{\boldsymbol{p}}= \left[ \begin{matrix}
	\sqrt{\lambda _1}&		0&		\cdots&		\,\, 0\\
	0&		\sqrt{\lambda _2}&		0&		\,\,\vdots\\
	\vdots&		0&		\ddots&		\,\, \   0\\
	0&		\cdots&		0&		\sqrt{\lambda _d}\\
\end{matrix},\hspace{0.2cm} 0_{d\times \left( 2n-d \right)} \right] Q^{-1}, 
$$
where $2n$ is the dimension of $X$ and $\lambda _1,\lambda _2,\cdots \lambda _d$ are the $d$ largest eigenvalues of $X$. This approach, however, unable to maintain the (primary) constraint imposed by constraint~(\ref{eq:z constraint A}) for the (1,2) block in $Z$. Since the relative measurements with respect to the z-axis are neglected, a natural consequence is that $\hat{\boldsymbol{p}}$ is unable to recover the correct 3D relative positions but only (when a certain regularization term is imposed \cite{biswas2008distributed}) the "shape" composed of multiple robots. 

Consequently, in the following sections we resort to other methods, focusing mainly on (i) the ability to infer the relative state efficiently and scalably, where distributed techniques become a natural choice, and (ii) the support for reasonably extracting $\boldsymbol{p}$.

\section{A Nonlinear Programming Model for Robotic Sensor Network Localization}
\label{Sec3}
In this section, we propose an unconstrained nonlinear programming (NLP) model that facilitates using the block coordinate descent (BCD)   algorithm for efficient solutions.  We follow the proposal of Burer and Monteiro \cite{burer2003nonlinear} to factorize $Z$ with some low-rank matrices $Y\in \mathbb{R} ^{ r\times \left( 2n+d \right)}$:
\begin{equation}
	\label{eq:B-M factorization}
	Z=Y^TY=\left[ \begin{array}{c}
		U^T\\
		Q^T\\
	\end{array} \right] \left[ \begin{matrix}
		U& Q\\
	\end{matrix} \right] ,\  Q^TQ=I_d,
\end{equation}
where $U\in \mathbb{R} ^{r\times 2n}$ and $Q\in \mathbb{R} ^{r\times d}$ with $r$ satisfying \mbox{$d\leq r\leq 2n+d$}, being the rank upper bound we impose on $Z$. This directly produces an NLP model that (i) implies a constraint on the rank of $Z$ rather than ignoring it, (ii) has a significantly lower dimension of the optimization variable relative to $Z$ as long as we take $r\ll n$, and (iii) leaves the explicit positive semidefiniteness constraint redundant since $Y^TY\succeq 0$, which is shown as follows:
\begin{equation}
	\label{eq:cost 1}
	\hspace{-0.2cm}\underset{U,Q}{\min}\hspace{-0.3cm}\sum_{\footnotesize\begin{array}{c}
			\left( i,j \right) \in \mathcal{E}\\
			u,v\in \mathcal{B}\\
	\end{array}}\hspace{-0.3cm}\frac{1}{ (\sigma _{ij}^{uv})^2 }{\left( \mathcal{A} _{ij}^{uv}\cdot \left[ \begin{array}{c}
			U^T\\
			Q^T\\
		\end{array} \right] \left[ \begin{matrix}
			U&		Q\\
		\end{matrix} \right] -\tilde{q}_{ij}^{uv} \right) ^2},
\end{equation}
\vspace{-0.7cm}
\begin{equation}
	\label{eq:O(d) constraint}
	Q^TQ=I_d,
\end{equation}
with constraints (\ref{eq:norm constraint A}), (\ref{eq:z constraint A}) where $Z$ is replaced with $U$, $Q$.  

Next, motivated by computational considerations, a more tractable model is built on top of (\ref{eq:cost 1}):  

\textbf{Fixing the projection operator:} As seen from the constitutive relation from $U$ and $Q$ to $Z$, $Q^TU$ is exactly the realization of $\boldsymbol{p}$. Therefore, $U$ can be considered a $r$-dimension surrogate of $\boldsymbol{p}$ and $Q$ a (row) projection operator that maps $U$ onto the realization space $\mathbb{R} ^d$.  

In the proposed NLP model, we choose a specific $Q$ and fix it throughout the optimization procedure\footnote{Although whether $Q$ is fixed or not has no effect on the optimal values of the model and the corresponding estimates, it does not mean that both nonlinear programming models will produce the same behavior and achieve the same estimate in the local search process. There is no theoretical evidence to compare the estimation accuracy of the two models. Still, the optimization procedure of the model with fixed $Q$ is expected more efficient since no preserving of orthogonality constraint \cite{jiang2015framework} is required.}. As a result, our model does not maintain the orthogonality constraint~(\ref{eq:O(d) constraint}). Note that with $Q$ fixed, different choices of $Q$ do not intrinsically affect the model's behavior, so we can trivially determine $Q$ as $\left[ \begin{matrix}
	I_d&		{0}\\
\end{matrix}_{d\times \left( r-d \right)} \right]^T$, for example.

\textbf{Approximating the equality constraints:} We propose to employ a quadratic penalty \mbox{method \cite[Chapter 17.1]{nocedal1999numerical}} to approximate the equality constraints, which yields an unconstrained NLP model.  

Specifically, we replace (\ref{eq:norm constraint A}) and (\ref{eq:z constraint A}) with the following terms, respectively:
\vspace{-0.25cm}
\begin{equation}
	\frac{1}{(\sigma _{i|\nu})^2}\left( \mathcal{A} _{ii}^{uv}\cdot \left[ \begin{array}{c}
		U^T\\
		Q^T\\
	\end{array} \right] \left[ \begin{matrix}
		U&		Q\\
	\end{matrix} \right] -d\nu _i \right) ^2,
\end{equation}
\vspace{-0.2cm}
\begin{equation}
	\label{eq:z penalty}
	\frac{1}{(\sigma _{i|z})^2}\left( \mathcal{A} _{i|z}\cdot \left[ \begin{array}{c}
		U^T\\
		Q^T\\
	\end{array} \right] \left[ \begin{matrix}
		U&		Q\\
	\end{matrix} \right] -dz_i \right) ^2,
\end{equation}
where \vspace{-0.15cm}
\begin{equation}
	d\nu _i=\left\| \bar{\nu}_{i}^{u}-\bar{\nu}_{i}^{v} \right\| ^2,
\end{equation}
\vspace{-0.4cm}
\begin{equation}
	\label{eq:dz}
	\hspace{-0.1cm}dz_i\hspace{-0.08cm}=\hspace{-0.08cm}\left[ \begin{matrix}
		-\sin \tilde{\phi}_i,\	\cos \tilde{\phi}_i\sin \tilde{\theta}_i, \		\cos \tilde{\phi}_i\cos \tilde{\theta}_i\\
	\end{matrix} \right] \hspace{-0.1cm}\left( \bar{\nu}_{i}^{u}-\bar{\nu}_{j}^{v} \right),
\vspace{-0.1cm}
\end{equation}
and the penalty coefficients $\sigma _{i|\nu}$ and $\sigma _{i|z}$ are parameters set to be small relative to the noise level $\sigma _{ij}^{uv}$.  

\textbf{Introducing difference into the factors:} A recent work~\cite{xu2013block} provides strong theoretical guarantees for solving \emph{block multiconvex} \cite[Sec. 1]{xu2013block} problems with the cyclic BCD algorithm. With these findings, authors in \cite{nishijima2022block} demonstrated how to construct the block multiconvex problem by introducing \mbox{\emph{difference}} into the factors $Y^T$ and $Y$ in the BM factorization, and take advantages of such difference in optimization. In this way, factorization~(\ref{eq:B-M factorization}) can be rewritten~as
\begin{equation}
	\label{difference B-M factorization}
	Z=Y^TY=\left[ \begin{array}{c}
		U^T\\
		Q^T\\
	\end{array} \right] \left[ \begin{matrix}
		V&		Q\\
	\end{matrix} \right] ,\ U=V, \ Q^TQ=I_d.
\end{equation}
The difference constraint $U=V$ is added to the objective function as a penalty term $\gamma*\|U-V\|_{F}^2$. So far, we have obtained the following unconstrained NLP model the BCD algorithm finally works~on:
\vspace{0.1cm} \\
\textbf{Problem 4} (Unconstrained NLP model for robotic sensor network localization with factorization difference).
\begin{equation}
	\label{eq:cost 2}
	\begin{split}
		\hspace{-0.18cm}\underset{U\in \mathbb{R} ^{r\times 2n}}{\min}\hspace{-0.2cm}F(U,V)=\hspace{-0.5cm}\sum_{\footnotesize \begin{array}{c}
				\left( i,j \right) \in \mathcal{E}\\
				u,v\in \mathcal{B}\\
		\end{array}}\hspace{-0.3cm}{\frac{1}{(\sigma _{ij}^{uv})^2}\hspace{-0.05cm}\left(\hspace{-0.05cm} \mathcal{A} _{ij}^{uv}\cdot \left[ \begin{array}{c}
				U^T\\
				Q^T\\
			\end{array} \right]\hspace{-0.1cm} \left[ \begin{matrix}
				V&	\hspace{-0.1cm}	Q\\
			\end{matrix} \right] -\tilde{q}_{ij}^{uv} \hspace{-0.05cm}\right) ^2} \\ + \sum_{i\in \mathcal{V}}{\frac{1}{(\sigma _{i|\nu})^2}\hspace{-0.05cm}\left(\hspace{-0.05cm} \mathcal{A} _{ii}^{uv}\cdot \left[ \begin{array}{c}
				U^T\\
				Q^T\\
			\end{array} \right] \hspace{-0.1cm}\left[ \begin{matrix}
				V&	\hspace{-0.1cm}	Q\\
			\end{matrix} \right] -d\nu _i \hspace{-0.05cm}\right)^2} \\ + \sum_{i\in \mathcal{V}}{\frac{1}{2(\sigma _{i|z})^2}\hspace{-0.05cm}\left(\hspace{-0.05cm} \mathcal{A} _{i|z}\cdot \left[ \begin{array}{c}
				V^T\\
				Q^T\\
			\end{array} \right] \hspace{-0.1cm}\left[ \begin{matrix}
				U&	\hspace{-0.1cm}	Q\\
			\end{matrix} \right] -dz_i \hspace{-0.05cm}\right) ^2} \\
		+\sum_{i\in \mathcal{V}}{\frac{1}{2(\sigma _{i|z})^2}\hspace{-0.05cm}\left(\hspace{-0.05cm} \mathcal{A} _{i|z}\cdot \left[ \begin{array}{c}
				U^T\\
				Q^T\\
			\end{array} \right] \hspace{-0.1cm}\left[ \begin{matrix}
				V&	\hspace{-0.1cm}	Q\\
			\end{matrix} \right] -dz_i \hspace{-0.1cm}\right) ^2}
		+\gamma \left\| U-V \right\| _{F}^{2}.
	\end{split}
\end{equation}

We summarize the main motivations of introducing such a difference as follows:

(i) The form of $F$ allows obtaining a closed-form solution (Proposition 1) efficiently for each \emph{block update subproblem} in the BCD framework when a reasonable block division (as will be explained in Section \ref{sec: 5A}) is applied. 

(ii) Under reasonable block division, each block update subproblem is strongly convex to meet Assumption~2 of~\cite{xu2013block}. Therefore, some good convergence properties proposed in~\cite{xu2013block} can be established for the proposed BCD algorithm (Theorem~1).

\section{A Decomposable Convex  Formulation for Robotic Sensor Network Localization}
While Problem 4 is expected to be solved efficiently using NLP algorithms, it require a (more or less) reasonable initialization. Motivated by this, we propose a convex formulation for Problem 3 that gives informed estimates regardless of the initial guess. 

In particular, in additional to relaxing the rank constraint in (\ref{eq:Z>0}), we futher relax $Z\succeq 0$ as follows~\cite{wang2008further}:
\begin{equation}
	\label{eq:ESDR}
	\left[ \begin{matrix}
		X_{2i+u,2i+u}&		X_{2i+u,2j+v}&		\left( p_{i}^{u} \right) ^T\\
		X_{2j+v,2i+u}&		X_{2j+v,2j+v}&		\left( p_{j}^{v} \right) ^T\\
		p_{i}^{u}&		p_{j}^{v}&		I_d\\
	\end{matrix} \right] \succeq 0, \forall \left( i,j \right) \in \mathcal{E} \lor u\ne v.
\end{equation}
We note that these constraints only appear between nodes where measurement edges exist in $\mathcal{G}$ or between variables corresponding to multiple sensors on the one robot, hence the result problem is called edge-based SDP (ESDP). This relaxation of the semidefinite constraint allows the problem to be decomposable and solved in a distributed mannner.

\textbf{Anchored ESDP:} Unfortunately, in ESDP literatures \cite{wang2008further,pong2011robust}, anchors are configured to be uniformly and sufficiently distributed, which is impractical in our problem, otherwise ESDP would not work well. To this end, our key observation is that in the case where, despite the presence of only a \emph{few} (for example, $<$5\% of the number of robots) imperfectly localized anchors, it is possible to recover informed estimates via ESDP if these anchor sensors generate measurements with a sufficient proportion of other (albeit non-neighboring) nodes. \vspace{0.1cm} \\ 
\emph{Remark 1} (Practicality of anchored measurements). Existing techniques (e.g., \cite{nguyen2023relative}) can estimate coarse relative states for a few robots, which means that in practice we can use these techniques to provide anchor information for a large-scale ESDP. Moreover, the requirement for sufficient "anchor-other" measurements is not unrealistic. For example, UWBs have the capacity of tens to hundreds tags and can also achieve decimeter precision even at ranges of more than 100 meters~\cite{nooploop}. We can achieve estimates for most of the robots without relatively expensive odometry by equipping only a few robots (anchors) with higher-power UWBs as well as short-term odometeries just for motion planning and control.\vspace{0.1cm}

Suppose we fix $n_a$ robots as anchors. We define a new node set $\mathcal{N} ^{\prime}=\mathcal{N} /\mathcal{N} _a$ where $\mathcal{N} _a$ is the anchor set, and measurement topologies $\mathcal{E} ^{\prime}$  and $
\mathcal{G} ^{\prime}$ corresponding to $\mathcal{N} ^{\prime}$. We also define separately the set $\mathcal{E} _a$ representing measurement edges between anchors and other nodes. As before, we can still express the measurment relationship between node $(i,u)$ and anchor $(k,v)$
\begin{equation}
	\begin{split}
			\left\| p_{i}^{u}-\hat{a}_{k}^{v} \right\| ^2=Z_{2i-1+u,2i-1+u}-2\left( \hat{a}_{k}^{v} \right) ^Tp_{i}^{u}+\left( \hat{a}_{k}^{v} \right)^T\hat{a}_{k}^{v}
	\end{split}
\end{equation}
with a linear mapping $\left\| p_{i}^{u}-\hat{a}_{k}^{v} \right\| ^2=\mathcal{A} _{ik}^{uv}\cdot Z$, where $\hat{a}_{k}^{v}$ is the (imperfect) estimate of anchor $(k,v)$. Then we obtain the following ESDP problem with anchor information: \vspace{0.1cm} \\
\textbf{Problem 5} (Anchored edged-based relaxation for robotic sensor network localization).
\begin{equation}
	\begin{split}
		\underset{Z}{\min} \ G(Z)=\hspace{-0.5cm}\sum_{\footnotesize{\begin{array}{c}
					\left( i,j \right) \in \mathcal{E}\\
					u,v\in \mathcal{B}\\
		\end{array}}}{\frac{1}{(\sigma _{ij}^{uv})^2}}\left( \mathcal{A} _{ij}^{uv}\cdot Z-\tilde{q}_{ij}^{uv} \right) ^2+ \\ \sum_{\footnotesize{\begin{array}{c}
					\left( i,k \right) \in \mathcal{E} _a\\
					u,v\in \mathcal{B}\\
		\end{array}}}{\frac{1}{(\sigma _{ik}^{uv})^2}}\left( \mathcal{A} _{ik}^{uv}\cdot Z-\tilde{q}_{ik}^{uv} \right) ^2
	\end{split}
	\vspace{-1.2cm}
\end{equation}
\vspace{-0.5cm}  s.t. 
\begin{center}
$$
	\left[ \begin{matrix}
		X_{2i+u,2i+u}&		X_{2i+u,2j+v}&		\left( p_{i}^{u} \right) ^T\\
		X_{2j+v,2i+u}&		X_{2j+v,2j+v}&		\left( p_{j}^{v} \right) ^T\\
		p_{i}^{u}&		p_{j}^{v}&		I_d\\
	\end{matrix} \right] \succeq 0, \forall \left( i,j \right) \in \mathcal{E} ^{\prime},
$$
$$
\left[ \begin{matrix}
	X_{2i+u,2i+u}&		X_{2i+u,2j+v}&		\left( p_{i}^{u} \right) ^T\\
	X_{2j+v,2i+u}&		\left( a_{k}^{v} \right) ^Ta_{k}^{v}&		\left( a_{k}^{v} \right) ^T\\
	p_{i}^{u}&		a_{k}^{v}&		I_d\\
\end{matrix} \right] \succeq 0,\forall \left( i,k \right) \in \mathcal{E} _a,
$$

\end{center}
$$
	\mathcal{A} _{ii}^{uv}\cdot Z=\left\| \bar{\nu}_{i}^{u}-\bar{\nu}_{i}^{v} \right\|^2, \ \forall i\in \mathcal{V} , u,v\in \mathcal{B} , u< v,
$$
$$
	\mathcal{A} _{i|z}\cdot Z=\left[ \begin{matrix}
		-\sin \tilde{\phi}_i,\	\cos \tilde{\phi}_i\sin \tilde{\theta}_i, \		\cos \tilde{\phi}_i\cos \tilde{\theta}_i \\
	\end{matrix} \right] \left( \bar{\nu}_{i}^{u}-\bar{\nu}_{j}^{v} \right), \vspace{-0.1cm}
$$
$$
\forall u,v\in \mathcal{B} , u< v.
$$

\section{Distributed Optimization in Robotic Seneor Network with Block Coordinate Descent}
\label{Sec 5}
For distributed implementations, BCD methods has been used for SNL~\cite{pong2011robust} as well as for multi-robot SLAM~\cite{tian2021distributed}. Inspired by these efforts, we develop algorithms with the classical BCD method as their backbone. 

As preparation, we first recap the BCD framework, which optimizes the following problem
\begin{equation}
	\min_{\boldsymbol{x}\in \mathcal{X}} f\left( \boldsymbol{x} \right),
\end{equation}
where $\mathcal{X}$ is the feasible set and assumed closed. Letting \emph{block division}  $\{\varPhi _q\}_{q=1,...,p}: \boldsymbol{x}=\left( x_1,...,x_q \right) $ divide $\boldsymbol{x}$ into $q$ disjoint blocks, the BCD algorithm of Gauss-Seidel type \cite{bertsekas2015parallel} optimizes the following \emph{block update subproblem} over each block $\varPhi _q$ cyclically 
\vspace{-0.3cm} 
\begin{equation}
	\label{eq:block update problem}
	\min_{x_q\in \mathcal{X} _q} f\left( \hat{x}_1,...,x_q,...,\hat{x}_p \right),
	\vspace{-0.3cm} 
\end{equation}
where $\hat{x}_i$ is the (temporarily) fixed value of remaining blocks and $
\mathcal{X} _q\coloneqq \left\{ x_q:\left( \hat{x}_1,...,x_q,...,\hat{x}_p \right) \in \mathcal{X} \right\} $.

In the context of our problem, block division  $\{\varPhi _q\}$ is defined as disjoint subsets of the ID set $\mathbb{S} \coloneqq \left\{ \left( i,u \right) |\forall i\in \mathcal{E} , u\in \mathcal{B} \right\}$
that identifies the variables corresponding to a certain sensor, such that $\bigcup\nolimits_{q=1}^p{\varPhi _q}=\mathbb{S}$. We outline the BCD algorithm that \emph{exactly} solves each subproblem (\ref{eq:block update problem}) for solving Problem 4 and Problem 5 in Algorithm \ref{alg}. 
\begin{algorithm} 
	\caption{\emph{Distributed Block Coordinate Descent for Robotic Sensor Network Localization}} 
	\label{alg} 
	\begin{algorithmic}[1]
		\renewcommand{\algorithmicrequire}{\textbf{Input:}}
		\renewcommand{\algorithmicensure}{\textbf{Output:}}
		\REQUIRE Block division $\{\varPhi _q\}$, \mbox{an initialization guess $\boldsymbol{p}^{\left( 0 \right)}$}       
		\ENSURE Robots' states estimates $\{ ( \hat{R}_i,\hat{t}_i ) \} $
		\STATE $k \gets 0$
		\WHILE {1} 
		\STATE $k \gets k+1$
		\FOR {$q=1,...,p$}
		\STATE Exactly solve block update subproblems (\ref{eq:block update problem}) w.r.t. $\varPhi _q$  
		\STATE Each robot with sensor  $\left( i,u \right) \in 
		\varPhi _q$ transmit the just updated variables to all its neighbors 
		\ENDFOR
		\IF{${\left\| \boldsymbol{p}^{\left( k   \right)}-\boldsymbol{p}^{\left( k-1   \right)} \right\| _F}/{\left\| \boldsymbol{p}^{\left( k-1   \right)} \right\| _F} \leq \epsilon $} 
		\STATE \textbf{break}
		\ENDIF
		\ENDWHILE
		\STATE Refinement in Problem 4 where $r=d$ and $Q=I_d$ using block update in Proposition 1 
		\STATE Recover $\{ ( \hat{R}_i,\hat{t}_i ) \} $ from $\boldsymbol{p}^{\left( k   \right)}$ according to (\ref{eq:p=Rv+t}), (\ref{eq:sinpsi}) and~(\ref{eq:cospsi})
	\end{algorithmic} 
\end{algorithm}
  
\subsection{BM-BCD: Fast Local Search on Problem 4}
\label{sec: 5A}
\subsubsection{Main idea} 
We first present the main idea of the algorithmic design in BM-BCD. 

\textbf{Exact block update:} In BM-BCD, the block updates (for example, line 7 of Algorithm \ref{alg1}) are performed by solving~(\ref{eq:block update problem}) exactly. More precisely, the update subproblem w.r.t. variable $U_{i}^{u}$ which is the $2i-1+u$ column of $U$ (and is actually a surrogate for the realization of sensor $(i,u)$) for any $1\leq i\leq n$ and $u\in \left\{ 0,1 \right\}$, can be solved exactly in a closed-form. This result is formalized as follows:  \vspace{0.15cm} \\
\textbf{Proposition 1.} Fix other columns of $U$ as well as $V$ arbitrarily. The solution of the block update subproblem~(\ref{eq:block update problem}) of Problem 4 w.r.t. $U_{i}^{u}$ are 
$({U}_{i}^{u})^*\hspace{-0.1cm}=\hspace{-0.1cm}( A_{i|d}^{u}+A_{i|\nu}^{u}+A_{i|z}^{u}+A_{i|\gamma}^{u} ) ^{-1}( b_{i|d}^{u}+b_{i|\nu}^{u}+b_{i|z}^{u}+b_{i|\gamma}^{u} )$ and $
({U}_{i}^{u})^*\hspace{-0.1cm}=\hspace{-0.1cm}( A_{i|d}^{u}+A_{i|\nu}^{u}+A_{i|\gamma}^{u} ) ^{-1}( b_{i|d}^{u}+b_{i|\nu}^{u}+b_{i|\gamma}^{u} ) $ for 3D and 2D problems respectively, where 
\vspace{-0.2cm}
$$
A_{i|d}^{u}=\sum_{\footnotesize\begin{array}{c}
		j\in \mathcal{N} _i\\
		v\in \mathcal{B}\\
\end{array}}{\frac{1}{(\sigma _{ij}^{uv})^2}\left( V_{i}^{u}-V_{j}^{v} \right) \left( V_{i}^{u}-V_{j}^{v} \right) ^T},
$$
\vspace{-0.2cm}
$$
b_{i|d}^{u}=\sum_{\footnotesize\begin{array}{c}
		j\in \mathcal{N} _i\\
		v\in \mathcal{B}\\
\end{array}}{\frac{1}{(\sigma _{ij}^{uv})^2}\left[ \left( V_{i}^{u}-V_{j}^{v} \right) ^TU_{j}^{v}+\tilde{q}_{ij}^{uv} \right] \left( V_{i}^{u}-V_{j}^{v} \right)},
$$
\vspace{-0.2cm}
$$
A_{i|\nu}^{u}=\frac{1}{(\sigma _{i|\nu})^2}\left( V_{i}^{u}-V_{i}^{v} \right) \left( V_{i}^{u}-V_{i}^{v} \right) ^T, \ (v\hspace{-0.1cm}=\hspace{-0.1cm}1-u)
$$
\vspace{-0.2cm}
$$
b_{i|\nu}^{u}=\frac{1}{(\sigma _{i|\nu})^2}\left[ \left( V_{i}^{u}-V_{i}^{v} \right) ^TU_{i}^{v}+d\nu _i \right] \left( V_{i}^{u}-V_{i}^{v} \right), \hspace{-0.1cm}\ (v\hspace{-0.1cm}=\hspace{-0.1cm}1-u)
$$
\vspace{-0.2cm}
$$
A_{i|z}^{u}=\frac{1}{2(\sigma _{i|z})^2}Q\left[ \begin{matrix}
	0&		0&		0\\
	0&		0&		0\\
	0&		0&		1\\
\end{matrix} \right] Q^T,
$$
\vspace{-0.15cm}
$$
b_{i|z}^{u}\hspace{-0.15cm}=\hspace{-0.1cm}\frac{1}{2(\sigma _{i|z})^2}\hspace{-0.1cm}\left(\hspace{-0.1cm} \left[ \begin{array}{c}
	0\\
	0\\
	1\\
\end{array} \right]^T\hspace{-0.3cm}Q^TU_{i}^{v}\hspace{-0.1cm}+\hspace{-0.1cm}(v-u)dz_i \hspace{-0.1cm}\right)\hspace{-0.0cm} \hspace{-0.1cm}Q\hspace{-0.1cm}\left[ \begin{array}{c}
	0\\
	0\\
	1\\
\end{array} \right] \hspace{-0.1cm}\hspace{-0.0cm},(v\hspace{-0.1cm}=\hspace{-0.1cm}1-u)
$$
\vspace{-0.2cm}
$$
A_{i|\gamma}^{u}=\gamma I_d, \ b_{i|\gamma}^{u}=\gamma V_{i}^{u}.
$$
By swapping the positions of $U$ and $V$ in the above equations, it is direct to obtain the updates of columns in~$V$. \vspace{0.2cm}

According to Proposition 1, the \mbox{closed-form} update only requires solving a low-dimensional linear system (with its dimension being \mbox{$r\ll n$}), so the exact updates have superior or comparable efficiency to the approximate ones. Meanwhile, since exact updates achieve the largest reduction of the overall cost function, fewer iterations are required than those algorithms with approximate updates, suppose that the algorithm converges. The following results show that each iteration of continuation (lines~20-27) and the refinement step (line 30) of Algorithm \ref{alg1} do asymptotically progress to a stationary point. \vspace{0.2cm} \\
\textbf{Assumption 1.} Every sensor point is connected, directly or indirectly, to at least one sensor with a fixed coordinate.   \vspace{0.2cm} \\
\textbf{Theorem 1.} Fix $\sigma _{i|\nu}$, $\sigma _{i|z}$ and $\gamma$.  Let $\left( U^{\left( k \right)},V^{\left( k \right)} \right)$ be the $k$-th iteration generated by Algorithm 1 for solving Problem 4 and $\mathcal{N}$ the set of stationary points of $F$. Under Assumption 1 and \emph{some} block division ensuring strongly convex subproblems, we have that $
\lim_{k\rightarrow \infty} \,\,\mathrm{dist}\left( \left( U^{\left( k \right)},V^{\left( k \right)} \right) , \mathcal{N} \right)=0$.

\emph{Proof.} Assumption~1 implies that the level set $\{(U,V)\mid\overset{}{F(U,V;\gamma)}\leq\alpha\}$ of $F$ is bounded \cite[Proposition 1]{so2007theory}. Therefore, the boundness of the sequence $\left\{ \left( U^{\left( k \right)},V^{\left( k \right)} \right) \right\} $ and the existence of a Nash equilibrium can be guaranteed according to Remark 2.4 and \mbox{Remark~2.2} of \cite{xu2013block}, respectively. For every block to which the exact update\mbox{\cite[Eq.(1.3a)]{xu2013block}} is applied, the subproblem is strongly convex, e.g., with a constant positive module $\sum_i{\left( \sigma _{i|z} \right) ^{-2}}+2\gamma$ \cite[Eq.(2.1)]{xu2013block} under the column-wise block division mentioned immediately below so that the Assumption 2 of \cite{xu2013block} is satisfied. Then Corollary~2.4 of~\cite{xu2013block} directly shows that the sequence converges to a Nash point of $F$ in the sense of $
\lim_{k\rightarrow \infty} \,\,\mathrm{dist}\left( \left( U^{\left( k \right)},V^{\left( k \right)} \right) , \mathcal{N} ^{\prime}\right)=0$, where $\mathcal{N} ^{\prime}$ is the set of Nash points. The Nash point must be one of the stationary points of $F$ since there is no constraint on the feasible set of $U$ and $V$, as claimed in Remark 2.2 of~\cite{xu2013block}. \vspace{-0.2cm} \\

Note that in the context of our problem, Assumption~1 does not generally hold because the knowledge of sensors' absolute locations are not necessary for relative state estimation. Still, we can artificially fix the coordinates of two sensors on a particular robot (as performed in line~17 of Alogirthm 2) without losing generality but guaranteeing the convergence.

\textbf{Block division:} Proposition 1 and the proof of Theorem~1 demonstrate the necessity of a reasonable block division for strong convexity of subproblems of Problem 4 and to enable closed-form updates. To achieve this, a reasonable block division scheme is to split \emph{each column} of $U$ and $V$, which surrogates a single sensor, into a separate block. 

For \emph{parallel} execution, these column-wise blocks can be further combined in a natural framework called graph coloring~\cite[Chapter 1.2.4]{bertsekas2015parallel},\cite{tian2021distributed}. A coloring scheme is provided for the \emph{dependency graph} that captures the dependencies relationship between variables. When two variables are independent of each other in their respective block update subproblems (in the context of this paper, no constraints and measurements between the two sensors surrogated by these variables), they can be assigned the same color and enter the same block. Remarkably, there are distributed algorithms available that can greedily identify a $\left( \varDelta +1 \right)$-colors scheme for an arbitrary graph \cite{barenboim2013distributed}, where $\varDelta$ denotes the maximum degree of the graph. As the \mbox{multi-robot} system scales up, $\varDelta$ can be controlled to remain nearly constant. In this case, the number of blocks is almost invariant to the scale of the system.

\textbf{Refinement:} Local refinement has been widely used in SNL and molecular conformation problems: starting from the estimate given by the upstream algorithm, the refinement stage uses a standard centralized nonlinear optimization method to find a stationary point of the original problem. Suppose that $r$ is chosen to be exactly the dimension $d$ in (\ref{eq:B-M factorization}), $U$ (or $V$) in Problem 4 itself can be considered a realization in $\mathbb{R} ^d$. Accordingly, we propose to emulate a distributed refinement step by setting $r=d$ as performed in line 30 of Algorithm 2.      

\subsubsection{Implementation details}
We propose several computational components to make BM-BCD work better, including the choice and maintance of penalty coefficients, etc.

\textbf{Continuation iterations:} An imitation of \emph{continuation}~\cite{hale2008fixed} is designed in BM-BCD compared to the vanilla BCD algorithm. This mechanism is implemented in the Function 2. It follows that the mechanism serves to iteratively improve the feasibility of the solution found by the approximate NLP model w.r.t. the equality constraints (\ref{eq:norm constraint A}) and (\ref{eq:z constraint A}). We note that fixing  $\mu _l$ and $\mu _z$ to very small values throughout the algorithm makes updating difficult (imagine that only the penalty terms come into play during the optimization process), and in this case, the algorithm finally terminates in the vicinity of the initialization.\vspace{0.1cm} \\
\emph{Remark 2} (Penalty coefficients of BM-BCD). The initial penalty coefficients $\sigma _{i|\nu}^{(1)}$, $\sigma _{i|z}^{(1)}$ have to be chosen judiciously. Here we provide a method of calculating these coefficients, which works well in practice: \vspace{-0.2cm}
\begin{equation}
	\sigma _{i|\nu}^{(1)}=0.2 *d\nu _i,
	\vspace{-0.1cm}
\end{equation}
\begin{equation}
	\sigma _{i|z}^{(1)}=\left( \frac{\partial dz}{\partial \phi}\bigg|_{\tilde{\phi}_i}+\frac{\partial dz}{\partial \theta}\bigg|_{\tilde{\theta}_i} \right) *\frac{\pi}{45},
	\vspace{-0.1cm}
\end{equation}
where $dz(\phi,\theta)$ is a multivariate function constructed by replacing $\tilde{\phi}_i$ and $\tilde{\theta}_i$ in (\ref{eq:dz}) with variables $\phi$ and $\theta$, respectively. We also found that in the case that we take $\mu _l$ and $\mu _z$ as 1/10$\sim$1/50 (which is commonly used in continuation methods), the number of iterations $N_c$ set as 2$\sim$3 is sufficient to achieve an accurate estimate.
\vspace{0.1cm}

\textbf{Dynamic-$\gamma$ iterations:} The choice of $\gamma$ may have a significant impact on the number of iterations required for BM-BCD. Therefore, additional maintenance of $\gamma$ is necessary as the price for introducing difference.  We  adopt a simple method, following \cite{nishijima2022block}, to calculate (lines 3 and~18) and update (line 21) $\gamma$ during the first few iterations of the algorithm in a distributed manner, which is outlined in Function 1. In particular,  each robot maintains a penalty coefficient $\gamma_i$ and calculate the value of the objective function w.r.t. itself and its neighbors at the end of an iteration, according to which robot~$i$ updates $\gamma_i$.  

We note that since line 11 of Algorithm \ref{alg1} limits the reduction rate of $F$, the termination of dynamic-$\gamma$ iterations can be guaranteed by the lower boundedness of $F$. 

\begin{algorithm} 
	\caption{\emph{Burer-Monteiro factorization with Block Coordinate Descent (BM-BCD)}} 
	\label{alg1} 
	\begin{algorithmic}[1]
		\renewcommand{\algorithmicrequire}{\textbf{Input:}}
		\renewcommand{\algorithmicensure}{\textbf{Output:}}
		\REQUIRE Block division $\{\varPhi _q\}$, \mbox{initialization $U^{(0)}=V^{(0)}$}    
		\ENSURE Robots' states estimates $\{ ( \hat{R}_i,\hat{t}_i ) \} $
		\STATE $k \gets 0$
		\STATE \textcolor{magenta}{// Dynamic-$\gamma$ iterations} 
		\STATE calculate $\gamma^{(1)}$ 
		\WHILE {1} 
		\STATE $k \gets k+1$
		\FOR {$q=1,...,p$}
		\STATE Exactly solve block update using Proposition 1 with corresponding constraints w.r.t. $\varPhi _q$  
		\STATE Each robot with sensor  $\left( i,u \right) \in 
		\varPhi _q$ updates $\gamma_i$ using Function 1
		\STATE Each robot with sensor  $\left( i,u \right) \in 
		\varPhi _q$ transmits the just updated variables to all its neighbors 
		\ENDFOR
		\IF{$\frac{F^{\left( k-1 \right)}-F^{\left( k \right)}}{F^{\left( k-1 \right)}}<\epsilon _F$ or \\  \hspace{-0.5cm}$\max\left\{\hspace{-0.1cm}{\frac{4\|U^{(k)}-V^{(k)}\|_F}{\|U^{(k)}\|_F+\|V^{(k)}\|_F},\frac{\|U^{(k)}-U^{(k-1)}\|_F}{\|U^{(k-1)}\|_F},\frac{\|V^{(k)}-V^{(k-1)}\|_F}{\|V^{(k-1)}\|_F}\hspace{-0.1cm}}\right\}\hspace{-0.1cm}\newline<\hspace{-0.1cm}\epsilon$}
		\STATE $(U^{(k)},V^{(k)}) \gets \left( \frac{U^{(k)}+V^{(k)}}{2},\frac{U^{(k)}+V^{(k)}}{2} \right) $
		\STATE Calculate $\gamma_{\rm final}$ and  fix $\gamma$ as $\gamma_{\rm final}$ 
		\STATE \textbf{break}
		\ENDIF
		\ENDWHILE
		\STATE Select $i$ and fix coordinates of sensors $(i,0)$ and~$(i,1)$
		\STATE \textcolor{magenta}{// Continuation iterations}
		\FOR{$N=1,...,N_c$} 
		\WHILE {1}
		\STATE \textbf{lines} 5-10
		\IF {$\max\left\{\frac{\|U^{(p)}-U^{(p-1)}\|_F}{\|U^{(p-1)}\|_F},\frac{\|V^{(p)}-V^{(p-1)}\|_F}{\|V^{(p-1)}\|_F}\right\}\hspace{-0.1cm}<\hspace{-0.1cm}\epsilon$}
		\STATE $(U^{(k)},V^{(k)}) \gets \left( \frac{U^{(k)}+V^{(k)}}{2},\frac{U^{(k)}+V^{(k)}}{2} \right) $
		\STATE $\sigma _{i|\nu}^{(N+1)} \gets \sqrt{\mu_l}*\sigma _{i|\nu}^{(N)}$, $\sigma _{i|z}^{(N+1)} \gets \sqrt{\mu_z}*\sigma _{i|z}^{(N)}$
		\STATE \textbf{break}
		\ENDIF
		\ENDWHILE
		\ENDFOR
		\STATE \textcolor{magenta}{// Refinement}
		\STATE Refine through line 2-28 with $r=d$ and $Q=I_d$ using block update in Proposition 1
		\STATE Recover $\{ ( \hat{R}_i,\hat{t}_i ) \} $ from $U^{(k)}$ according to (\ref{eq:p=Rv+t}), (\ref{eq:sinpsi}) and~(\ref{eq:cospsi})
	\end{algorithmic} 
\end{algorithm}

\subsection{ESDP-BCD: Exact Block Update on Problem 5} 
\textbf{Convex subproblems of ESDP:} We first show that the block update subproblem of Problem 5 w.r.t. robot $i$ (i.e., sensors $(i,0)$ and $(i,1)$) is also a convex programming.

Constraint (\ref{eq:ESDR}) can be rewritten as 
$$
	\left[ \begin{matrix}
		X_{2i+u,2i+u}-\left( p_{i}^{u} \right) ^Tp_{i}^{u}&		X_{2i+u,2j+v}-\left( p_{i}^{u} \right) ^Tp_{j}^{v}\\
		X_{2j+v,2i+u}-\left( p_{j}^{v} \right) ^Tp_{i}^{u}&		X_{2j+v,2j+v}-\left( p_{j}^{v} \right) ^Tp_{j}^{v}\\
	\end{matrix} \right] \succeq 0,
$$
of which all of the leading principal minors in the matrix are nonnegative, i.e.,
\begin{equation}
	X_{2i+u,2i+u}-\left( p_{i}^{u} \right) ^Tp_{i}^{u}\geqslant 0,
\end{equation}
\begin{equation}
	\label{eq:ESDR sub}
	\begin{split}
		\left| \begin{matrix}
			X_{2i+u,2i+u}-\left( p_{i}^{u} \right) ^Tp_{i}^{u}&		X_{2i+u,2j+v}-\left( p_{i}^{u} \right) ^Tp_{j}^{v}\\
			X_{2j+v,2i+u}-\left( p_{j}^{v} \right) ^Tp_{i}^{u}&		X_{2j+v,2j+v}-\left( p_{j}^{v} \right) ^Tp_{j}^{v}\\
		\end{matrix} \right|= \\ \left( X_{2i+u,2i+u}-\left( p_{i}^{u} \right) ^Tp_{i}^{u} \right) \left( X_{2j+v,2j+v}-\left( p_{j}^{v} \right) ^Tp_{j}^{v} \right) - \\ \left( X_{2i+u,2j+v}-\left( p_{i}^{u} \right) ^Tp_{j}^{v} \right) ^2\geqslant 0.
		\\
	\end{split}
\end{equation}
Fixing the components corresponding to the sensor on robot $j$ (e.g., sensor $(j,v)$ in (\ref{eq:ESDR sub})), we can directly obtain the the following block update subproblem of Problem 5 w.r.t. robot~$i$.
\begin{equation}
	\label{eq:convex subproblem}
	\begin{split}
		\underset{Z}{\min}\sum_{\footnotesize{\begin{array}{c}
					j\in \mathcal{N}_i 
					\\
					u,v\in \mathcal{B}\\
		\end{array}}}{\frac{1}{(\sigma _{ij}^{uv})^2}}\left( \mathcal{A} _{ij}^{uv}\cdot Z-\tilde{q}_{ij}^{uv} \right) ^2+  \\ \vspace{-1cm} \sum_{\footnotesize{\begin{array}{c}
					k\in \mathcal{C}_i\\
					u,v\in \mathcal{B}\\
		\end{array}}}{\frac{1}{(\sigma _{ik}^{uv})^2}}\left( \mathcal{A} _{ik}^{uv}\cdot Z-\tilde{q}_{ik}^{uv} \right) ^2
	\end{split}
	\vspace{-1.4cm}
\end{equation} 
\ \  \hspace{0.1cm} s.t.
\begin{equation}
	\label{eq:X>p2}
	X_{2i+u-1,2i+u-1} \geq \left\| p_{i}^{u} \right\| ^2, 
	\forall  u\in \mathcal{B} 	
	\vspace{-0.7cm}
\end{equation}
\begin{center}
	$$
	X_{2i+u-1,2i+u-1}\geq \left\| p_{i}^{u} \right\| ^2+\frac{1}{\hat{e}_{j}^{v}}\left( X_{2i+u-1,2i+u-1}-\left( p_{i}^{u} \right) ^Tp_{j}^{v} \right) ^2,
	\vspace{-0.2cm}
	$$
\end{center}
\begin{equation}
	\label{eq:edge constraint}
	\forall j\in \mathcal{N}_i, v\in \mathcal{B} ,
\end{equation}
$$
\mathcal{A} _{ii}^{uv}\cdot Z=\left\| \bar{\nu}_{i}^{u}-\bar{\nu}_{i}^{v} \right\|^2, u,v\in \mathcal{B} , u< v,
$$
$$
\mathcal{A} _{i|z}\cdot Z=\left[ \begin{matrix}
	-\sin \tilde{\phi}_i,\	\cos \tilde{\phi}_i\sin \tilde{\theta}_i, \		\cos \tilde{\phi}_i\cos \tilde{\theta}_i \\
\end{matrix} \right] \left( \bar{\nu}_{i}^{u}-\bar{\nu}_{j}^{v} \right), \vspace{-0.1cm}
$$
$$
u,v\in \mathcal{B} , u< v,
$$
where $\hat{e}_{j}^{v}$ is the (temporary) fixed value of the relaxation error $\left( X_{2j+v-1,2j+v-1}-\left( p_{j}^{v} \right) ^Tp_{j}^{y} \right)$ of sensor $(j,v)$ and $\mathcal{C}_l$ represents robots having measurements with anchor $l$. 

It is straightforward to see that the above problem is a (convex) quadratic conic programming that can be solved efficiently using off-the-shelf solvers.  The complexity of solving the problem simply depends on the robot $i$'s number of neighbors, just like the update approach shown in Proposition~1.\vspace{0.2cm} \\ 
\textbf{Theorem 2.} Let $Z^{(k)}$ be the $k$-th iteration generated by Algorithm 1 for solving Problem 5, where algorithms based on \emph{barrier function} are used to solve each subproblem, and $\mathcal{M}$ the set of global minimizers of $G$ under constraints. Under Assumption 1 and \emph{some} block division ensuring strongly convex subproblems, we have that $\lim_{k\rightarrow \infty} \,\,\mathrm{dist}\left( Z^{\left( k \right)}, \mathcal{M} \right) =0$.

\emph{Proof:} To prove Theorem 2, we first introduce the term of \emph{uncoupled constrained division}. If a optimization problem 
\begin{equation}
	\label{eq:constrained optimization problem}
	\min_X  G\left( X \right) \ \text{subjects to} \ H\left( X \right) =0
\end{equation}
can be rewritten as 
$$
	\min_X  G\left( X \right) \ \mathrm{subject} \ \mathrm{to} \  H_q\left( X_q \right) =0 \ \forall q=1,...,p
$$
under variable division $\{\varPhi _q\}_{q=1,...,p}: \ X=\left( X_1,...,X_p \right)$, the variable division is called an uncoupled constrained division of decision variable $X$. We show in the following Lemma that the Nash equilibrium under uncoupled constrained division also satisfies Karush-Kuhn-Tucker (KKT) conditions of the problem as long as $G\left( \cdot \right)$ and $H(\cdot)$ satisfying some common properties.\vspace{0.1cm} \\
\textbf{Lemma 1.} Let $G\left( \cdot \right)$, $H(\cdot)$ be once differential and (\ref{eq:constrained optimization problem}) satisfy strong duality. The Nash equilibrium of (\ref{eq:constrained optimization problem}) under uncoupled constrained division also satisfies the KKT conditions of (\ref{eq:constrained optimization problem}) under some constraint qualifications (e.g., the LICQ \cite[Chapter 12.2]{nocedal1999numerical}).  \vspace{0.1cm} \\
To prove this Lemma, we show that if $X^*=\left( X_{1}^{*},...,X_{p}^{*} \right)$ is a Nash point of (\ref{eq:constrained optimization problem}) under an uncoupled constrained division such that 
$$
G\left( X_{1}^{*},...,X_{q}^{*},...X_{p}^{*} \right) \leqslant G\left( X_{1}^{*},...,X_q,...X_{p}^{*} \right), \ \forall H_q\left( X_q \right) =0
$$
holds for all $q=1,...,p$, i.e., $X_{q}^{*}$ is the global minimizer of subproblem 
\begin{equation}
	\label{eq:subproblem}
	\min_{X_q} \,\, G( X_{1}^{*},...,X_{q},...X_{p}^{*}),\ \text{s.t.} \ H_q\left( X_q \right) =0.
\end{equation}
Due to the strong duality of (\ref{eq:constrained optimization problem}), it is straightforward to show that, as a global minimizer of subproblem (\ref{eq:subproblem}), $X_q^{*}$ is a KKT point of (\ref{eq:subproblem}) \cite[Chapter 5.5]{boyd2004convex}, i.e., there exists Lagrange multipliers $\lambda _{q}^{*}$ such that \cite[Theorem 12.1]{nocedal1999numerical}
\begin{equation}
	\label{Lagrangian}
	\nabla _{X_q}\mathcal{L} \left( X^*,\lambda _{q}^{*} \right) =\nabla _{X_q}G\left( X^* \right) +\lambda _{q}^{*}\nabla _{X_q}H_q\left( X_{q}^{*} \right) =0
\end{equation}
 $$\ H_q\left( X_q^{*} \right) =0,$$
where $\mathcal{L}_q(\cdot)$ is the Lagrangian function of (\ref{eq:subproblem}).  A straightforward computation shows that as we combine Eq. (\ref{Lagrangian}) over all $q$, we can obtain, since the division is uncoupled constrained, that
\begin{equation*}
	\begin{split}
			\nabla _X\mathcal{L} \left( X^*,\lambda ^* \right) =\nabla _XG\left( X^* \right) +\lambda ^*\nabla _XH\left( X^* \right) \\
		=[ \nabla _{X_1}G\left( X^* \right) +\lambda _{1}^{*}\nabla _{X_1}H_1\left( X_{1}^{*} \right) ;...; \\ \nabla _{X_p}G\left( X^* \right) +\lambda _{p}^{*}\nabla _{X_p}H_p\left( X_{p}^{*} \right) ] =0
	\end{split}
\end{equation*}
where $\lambda ^*=\left[ \lambda _{1}^{*},...,\lambda _{p}^{*} \right] $ 
and  $\mathcal{L}(\cdot)$ is the Lagrangian function of (\ref{eq:constrained optimization problem}). Similarly, $H(X^{*})=0$ natrually holds with (\ref{Lagrangian}) holding. Therefore, we conclude that the KKT conditions of (\ref{eq:constrained optimization problem}) is satisfied at $X^{*}$, so that Lemma 1 holds.

We claim that Lemma 1 can be directly applied to Problem 5 when the subproblem (\ref{eq:convex subproblem}) is solving using barrier function-based methods (e.g., interior point method), despite the inequality constraints (\ref{eq:ESDR}). This is because the inequality constraints will be replace to the objective function with a \emph{surrogate function} like the log-barrier function and a new objective function $G^{\prime}\left( Z \right)$ is obtained, with only equality constraints. It is not difficult to see that solving each subproblem (\ref{eq:convex subproblem}) exactly (ignoring the imperfect precision of solutions) using these methods is equivalent to solving $G^{\prime}\left( Z \right)$, with the remaining equality constraints, using exactly update, as long as the block division is uncoupled constarined. Moreover, it is easy to show that the subproblem satisfies Slater's condition \cite[Chapter 5.2]{boyd2004convex} so that the strong duality is guaranteed. Therefore, the assumptions of Lemma 1 hold.  

We can now prove Theorem 2 following the line of proof for Theorem 1. Similar to Theorem 1, the boundness of $Z^{(k)}$ and the exsitence of a Nash equilibrium can be guaranteed. For every block to which the exact update is applied, the subproblem is strongly convex, e.g., with a constant positive module $2\sum_{\footnotesize{\begin{array}{c}
			\left( i,j \right) \in \mathcal{E}\\
			u,v\in \mathcal{B}\\
\end{array}}}{(\sigma _{ij}^{uv})^{-2}} + 2\sum_{\footnotesize{\begin{array}{c}
\left( i,k \right) \in \mathcal{E} _a\\
u,v\in \mathcal{B}\\
\end{array}}}{(\sigma _{ik}^{uv})^{-2}}(a_{k}^v)^2$ under the block division by each single robot, i.e., the one implied by the subproblem (\ref{eq:convex subproblem}) so that the Assumption 2 of \cite{xu2013block} is satisfied.  Then Corollary~2.4 of~\cite{xu2013block} directly shows that the sequence converges to a Nash point of $G$ under constaints in the sense of $\lim_{k\rightarrow \infty} \,\,\mathrm{dist}\left( Z^{\left( k \right)}, \mathcal{M}^{\prime} \right) =0$, where $\mathcal{M} ^{\prime}$ is the set of Nash points. The Nash point must be one of the KKT points of Problem 5 according to Lemma 1 if some barrier function-based method is used to solve the subproblem. The KKT point must be one of the global minimizers due to the convexity of Problem 5. \vspace{0.15cm}

\textbf{Block division:} We note that the block update subproblem (\ref{eq:convex subproblem}) has implied a reasonable block division that split the variables corresponding to \emph{two sensors on one robot} (instead of splitting each sensor as in BM-BCD) into a separate block, which satisfies the assumptions of Theorem~2. Similar to the previous section, these blocks can be combined to achieve parallel execution using graph coloring.  

\textbf{Refinement:} Despite not relying on initial guesses, under noisy measurements, ESDP only provides coase estimates, especially for the rotation components of robots's state. For this reason, in our experiments we use the full Algorithm 2 as a refinement (which is slightly different from that described in Algorithm~1), since large numbers of $r$ will manifest higher robustness compared to fixing $r = d$ (see Section).

The full algorithm of ESDP-BCD is summarized in Algorithm 3.

\begin{algorithm} 
	\caption{\emph{Anchored Edge-based Semidefinite Programming with Block Coordinate Descent (ESDP-BCD)}} 
	\label{alg3} 
	\begin{algorithmic}[1]
		\renewcommand{\algorithmicrequire}{\textbf{Input:}}
		\renewcommand{\algorithmicensure}{\textbf{Output:}}
		\REQUIRE Block division $\{\varPhi _q\}$, \mbox{an initialization guess $\boldsymbol{p}^{\left( 0 \right)}$}, anchor set $\mathcal{N} _a$, anchor-other measurement topology $\mathcal{E} _a$     
		\ENSURE Robots' states estimates $\{ ( \hat{R}_i,\hat{t}_i ) \} $
		\STATE Each anchor robot transmit  an estimate of its state to all its neighbors
		\STATE $k \gets 0$
		\WHILE {1} 
		\STATE $k \gets k+1$
		\FOR {$q=1,...,p$}
		\STATE Exactly solve block update subproblem (\ref{eq:convex subproblem}) with corresponding constraints w.r.t. $\varPhi _q$  
		\STATE Each robot except anchor with sensor  $\left( i,u \right) \in 
		\varPhi _q$ transmits the just updated variables to all its neighbors 
		\ENDFOR
		\STATE Extract estimate $\boldsymbol{p}^{\left( k   \right)}$ from the (2,1) block of $Z^{(k)}$ in (\ref{eq:Z>0}) 
		\IF{${\left\| \boldsymbol{p}^{\left( k   \right)}-\boldsymbol{p}^{\left( k-1   \right)} \right\| _F}/{\left\| \boldsymbol{p}^{\left( k-1   \right)} \right\| _F} \leq \epsilon $} 
		\STATE \textbf{break}
		\ENDIF
		\ENDWHILE
		\STATE Refine using BM-BCD with some $r>d$ 
		\STATE Recover $\{ ( \hat{R}_i,\hat{t}_i ) \} $ from $\boldsymbol{p}^{\left( k   \right)}$ according to (\ref{eq:p=Rv+t}), (\ref{eq:sinpsi}) and~(\ref{eq:cospsi})
	\end{algorithmic} 
\end{algorithm}
 
\emph{Remark 3} (BCD algorithm: a natural choice for distributed settings and scalable optimization). To advance the BCD procedure, each robot $i$ only needs to solve the block update subproblems w.r.t. the sensors on it and communicate locally with its neighbors. Moreover, BCD can be easily executed in parallel and are suitable for asynchronous implementation, with the number of blocks depending on the maximum degree of $\mathcal{G}$ (recall section \ref{sec: 5A}). This fact implies that, ideally, the time taken for optimization and the communication overhead \vspace{0.1cm} per iteration are somewhat independent of the system's scale.  
\emph{Remark 4} (Exact block update: a communication-friendly algorithmic design for distributed settings). Our proposed methods that solves each block update subproblem exactly takes few iterations to converge. This alleviates the potential communication fragility of distributed architectures as it implies few rounds of communications.

\section{Numerical Experiments}
\vspace{-0.1cm}
\label{Sec 6}
In this section, we assess the performance of the proposed methods compared to alternative ones. Our simulation consists of four problems, where the robots respectively form a 3D cube, a 3D pyramid, a 2D hexagon and a 2D rectangle. See Table 1 for a overview of the problems, where $b$ is the ground-truth  distance between neighboring robots and the default noise level is $\sigma=0.1\mathrm{m}$. The positions of sensors on each robot are uniformly set as $\bar{\nu}_{i}^{0}=\left[ 0;0.35\mathrm{m};0 \right]$ and $\bar{\nu}_{i}^{1}=\left[ 0;-0.35\mathrm{m};0 \right] $ for $d=3$ or $\bar{\nu}_{i}^{0}=\left[ 0;0.35\mathrm{m}\right]$ and $\bar{\nu}_{i}^{1}=\left[ 0;-0.35\mathrm{m}\right] $ for $d=2$. We randomly generate each robot's initial location on a ball (for 3D problem) or circle (for 2D problem) with radius $\mathfrak{r}$ and  the ground-truth as its center and randomly generate the yaw angle of each robot. A feature value $
\rho =\frac{\mathfrak{r}}{\left( l-1 \right)b}$ is introduced to indicate the quality of initialization. 

We compare the proposed method with (i) \emph{Quasi-Newton} (QN), a standard central quadratic first-order algorithm for solving Problem 4 with fixed $r = d  $, (ii) \emph{Block Coordinate Gradient Descent} (BCGD), a BCD-type algorithm which replaces all the block updating steps in BM-BCD with a backtracking-based gradient descent method with $r = d+1$, and (iii) \emph{Anchored Semidefinite Programming} (SDP): a direct rank-relaxed version of Problem 3 that enjoys the same anchor information as ESDP-BCD. We solve each quadratic cone programming of ESDP-BCD as well as SDP using the interior point method implemented in MOSEK solver \cite{aps2019mosek}. QN and the MOSEK solver are both driven with 6 threads.

The termination conditions $\epsilon$ in Algorithm 1 are uniformly taken as $5\times 10^{-4}$ for NLP	methods and $5\times 10^{-2}$ for ESDP-BCD in the following experiments. The reason we choose a loose termination condition for the latters is that they do not provide highly accurate estimates and therefore does not need to be "fully convergent" before refinement. 

\begin{table}[]
	\hspace{-0.4cm}\label{table1}
	\setlength{\tabcolsep}{2.4pt}
	\renewcommand\arraystretch{1.2}
	\centering
	\begin{tabular}{cccc}
		\hline	Problem  & \hspace{-0.2cm}number of robots  & $b$(m) & max degree / min degree  \\ \hline
		\footnotesize\textbf{CUBE} (Fig. 2(a))	&  \hspace{-0.2cm}125 & 3 & \hspace{-0.2cm} 26 / 7   \\
		\footnotesize\textbf{PYRAMID} (Fig. 2(c)) &  \hspace{-0.2cm}84 & 4 & \hspace{-0.2cm} 24 / 6 \\
		\footnotesize\textbf{HEXAGON} (Fig. 2(e)) &  \hspace{-0.2cm}217 & 4.5 & \hspace{-0.2cm} 12 / 5  \\
		\footnotesize\textbf{RECTANGLE} (Fig. 2(g)) &  \hspace{-0.2cm}200 & 3 & \hspace{-0.1cm} 8 / 5  \\ \hline
		\vspace{0.01mm}
	\end{tabular}
	\textbf{Table 1.} Information of the 4 simulation problems. \vspace{-0.5cm}
\end{table}

\begin{figure*} \centering
	\hspace{-0.15cm}\subfigure[Failure rate FR and iterations k with different settings of $r$ in \textbf{PYRAMID}] {
		\includegraphics[width=1.00\columnwidth]{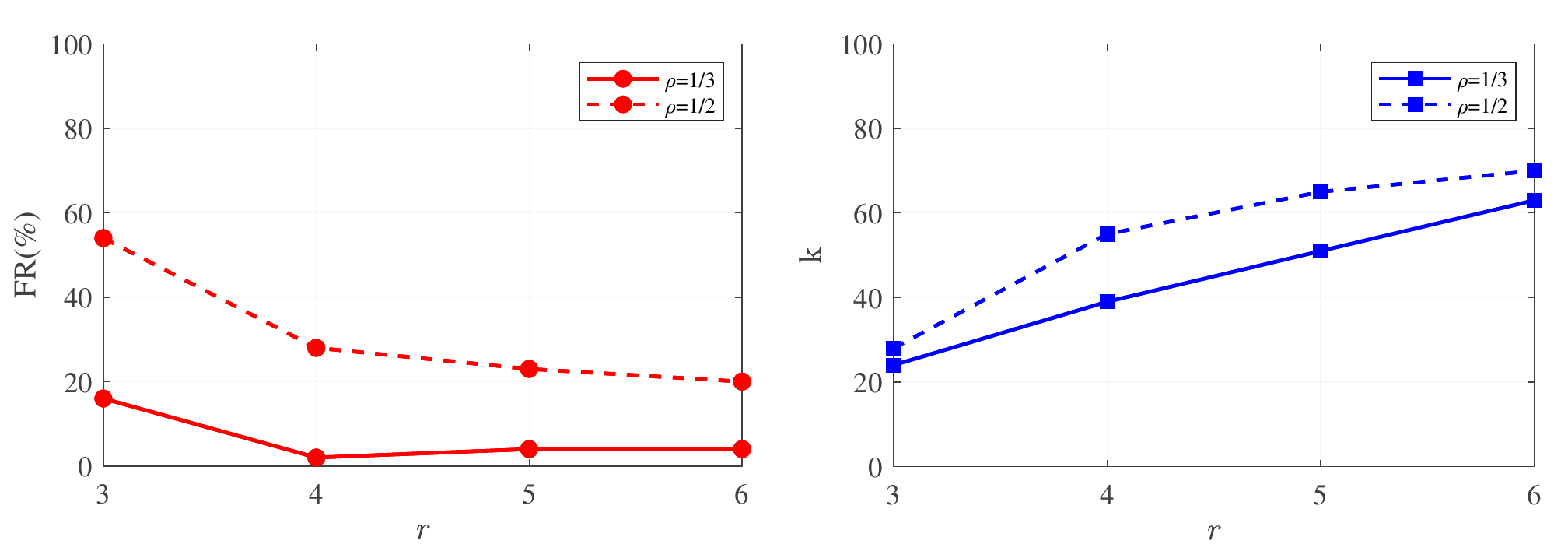}
	}      
	\hspace{-0.1cm}\subfigure[Failure rate FR and iterations k with different settings of $r$ in \textbf{RECTANGLE}] {    
		\includegraphics[width=1.00    \columnwidth]{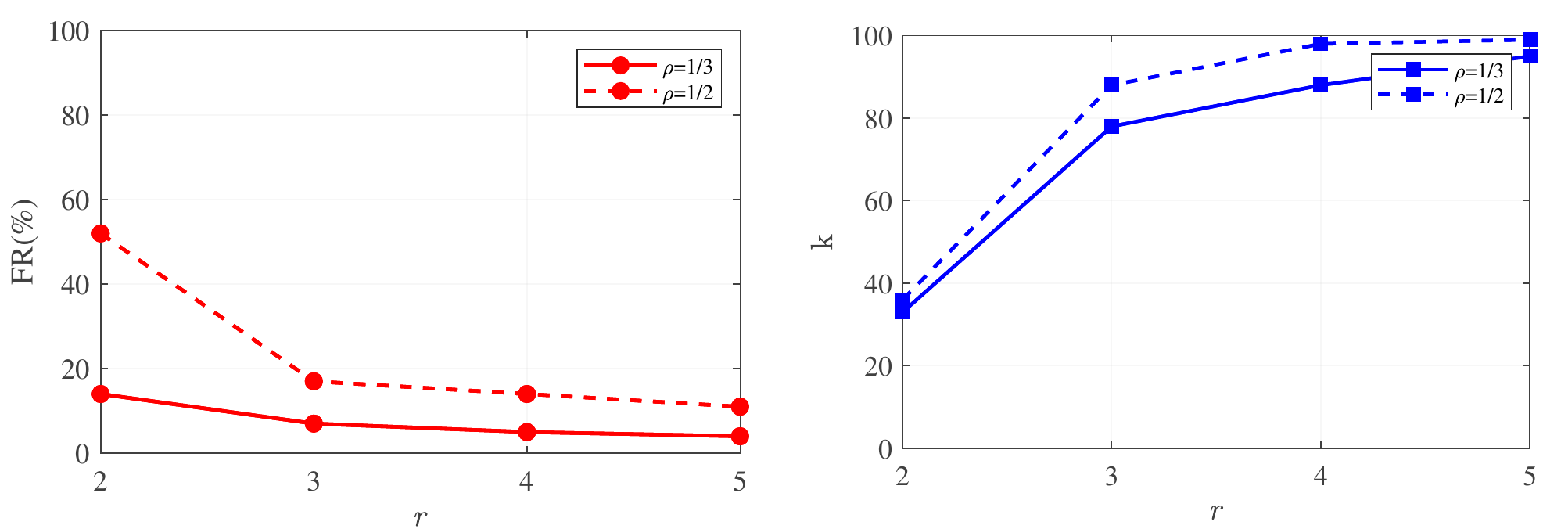}     
	} \\  
	\footnotesize\textbf{Figure 1.} Parametric evaluating of $r$ in BM-BCD on 2D and 3D problems.   
\end{figure*}

\begin{figure*} \centering
	\subfigure {  
		\includegraphics[width=0.58\columnwidth]{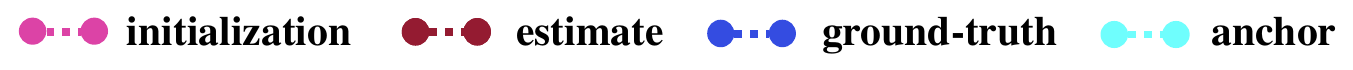}  
	} \vspace{-0.3cm}\\   
	\setcounter{subfigure}{0} 
	\hspace{-0.15cm}\subfigure[Initialization and result of BM-BCD($d$+1) in \textbf{CUBE}] {
		\label{fig:a2}     
		\includegraphics[width=1.00\columnwidth]{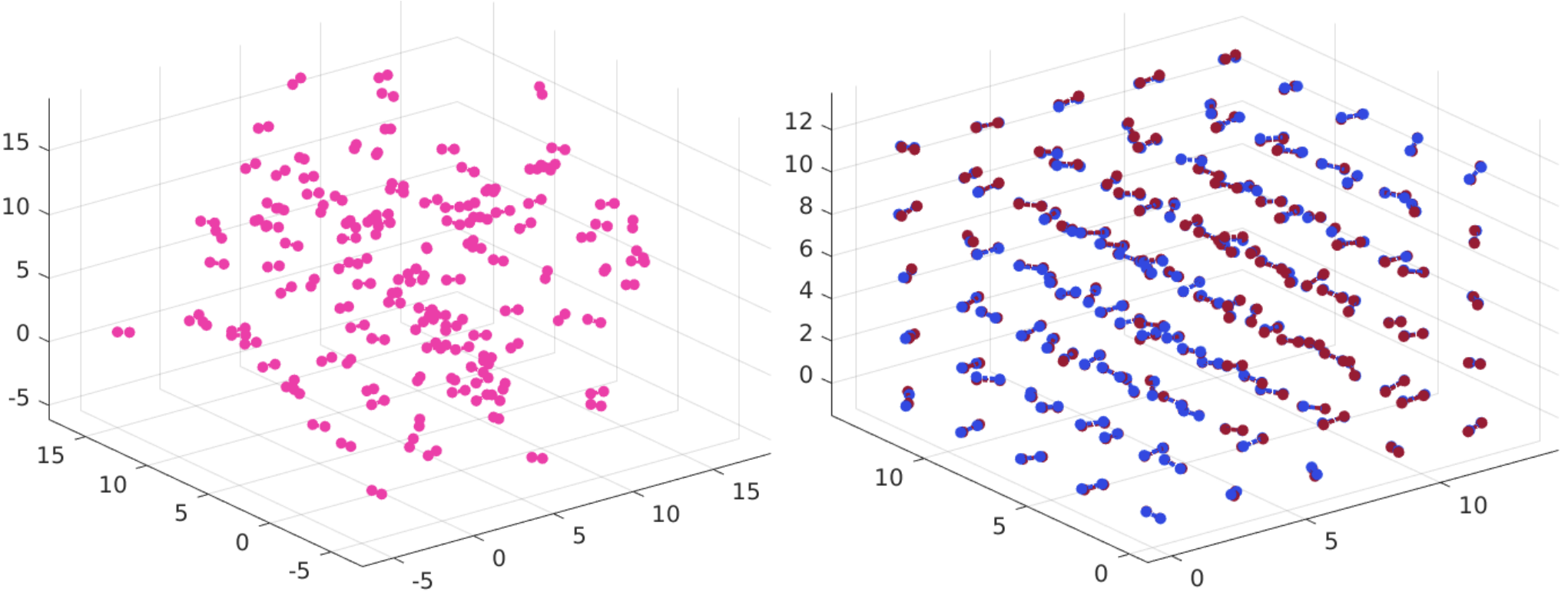}  
	}      
	\hspace{-0.3cm}\subfigure[Result before (left) and after (right) refinement of ESDP-BCD in \textbf{CUBE}] { 
		\label{fig:b2}     
		\includegraphics[width=1.02    \columnwidth]{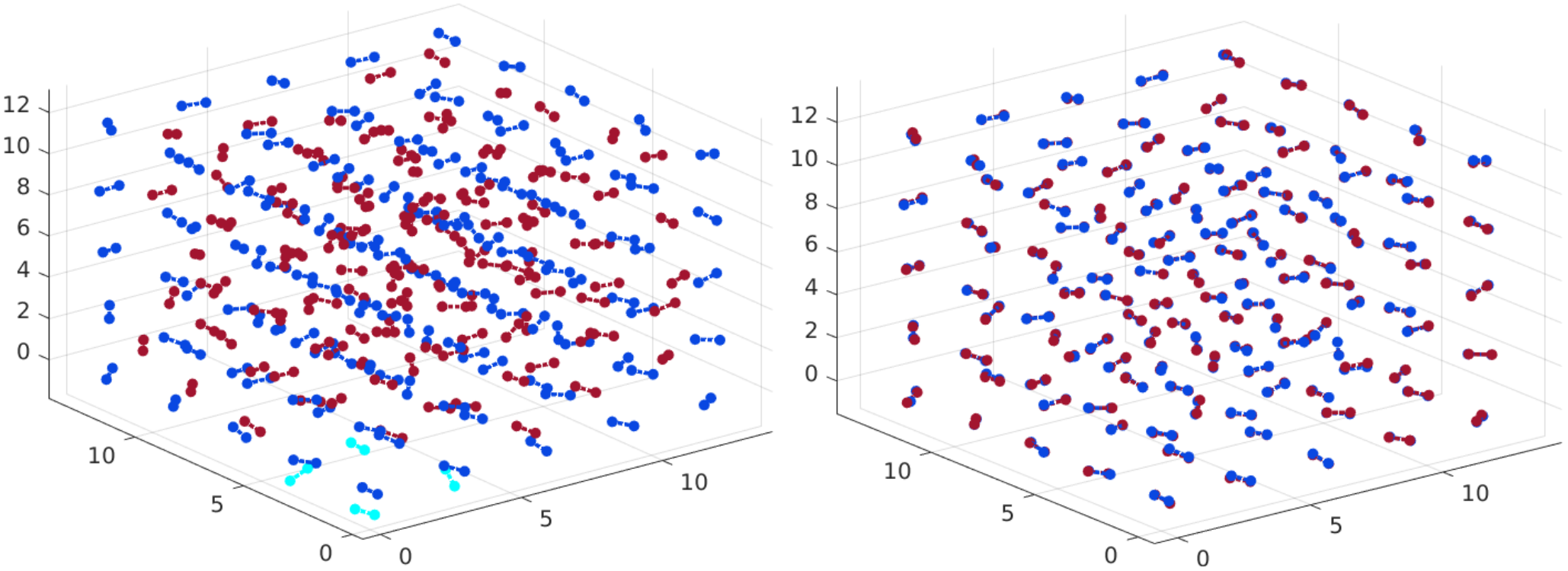}     
	} \\ \vspace{-0.2cm}
	\hspace{-0.15cm}\subfigure[Initialization and result of BM-BCD($d$+1) in \textbf{HEXAGON}] {
		\label{fig:c2}     
		\includegraphics[width=1.0\columnwidth]{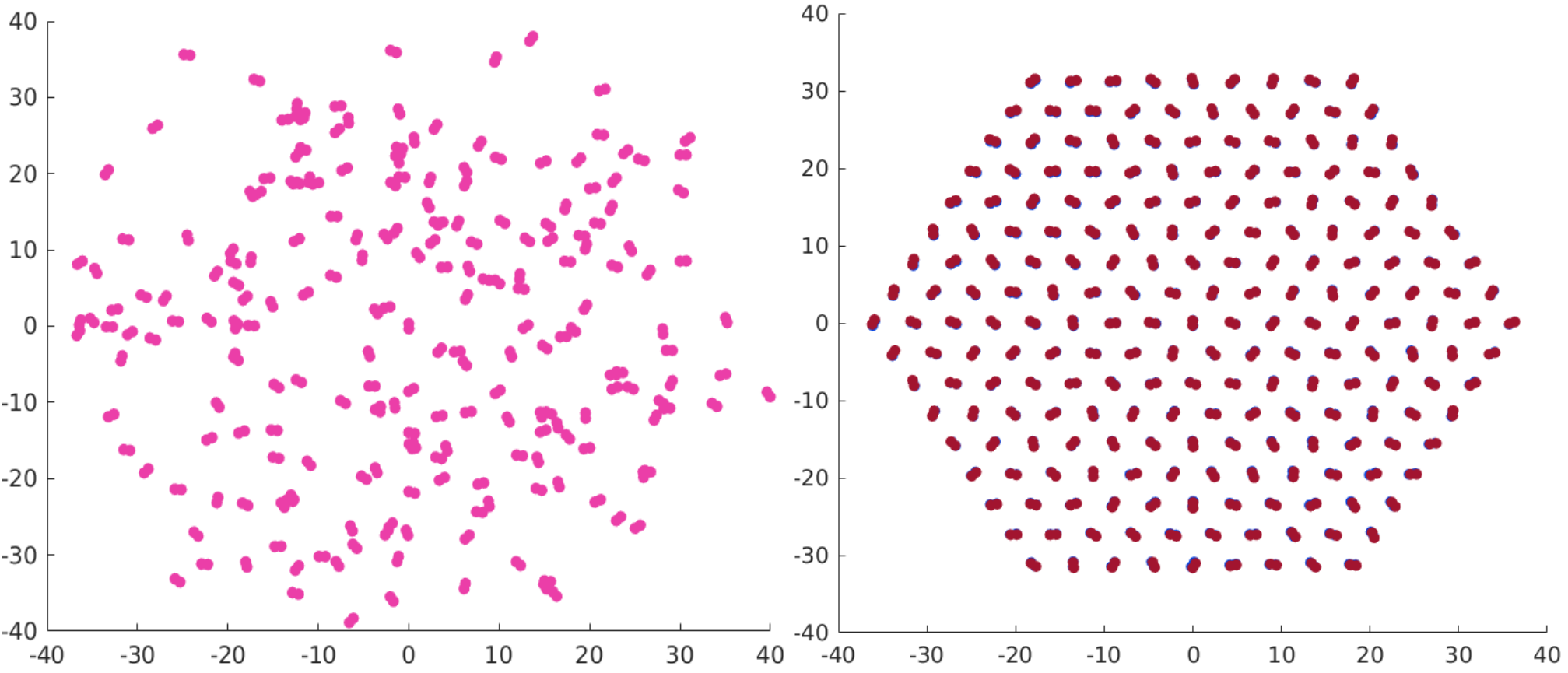}  
	}     
	\hspace{-0.25cm}\subfigure[Result before (left) and after (right) refinement of ESDP-BCD in \textbf{HEXAGON}] { 
		\label{fig:d2}     
		\includegraphics[width=0.95\columnwidth]{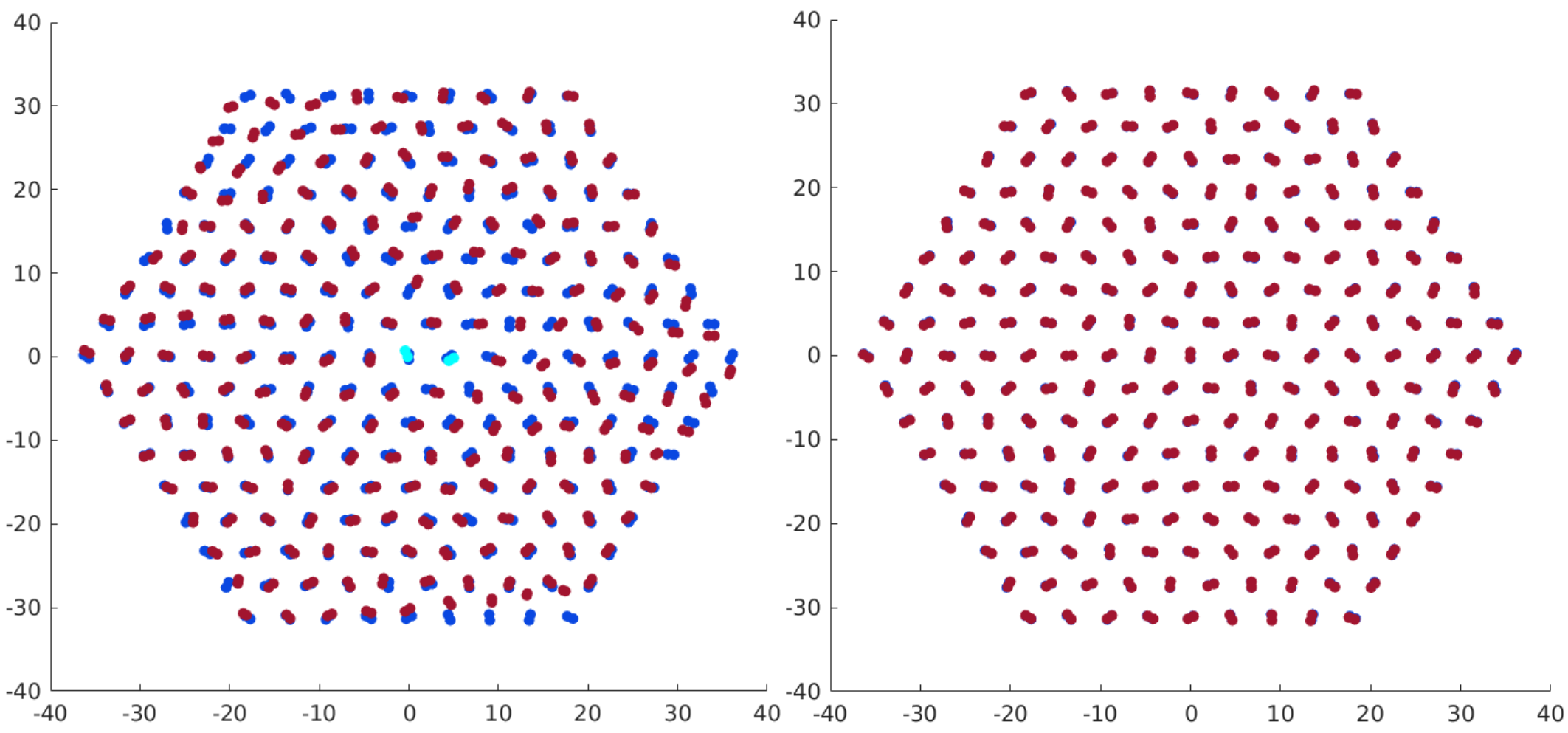}     
	}
	\hspace{-0.15cm}\subfigure[Initialization and result of BM-BCD($d$+1) in \textbf{PYRAMID}] {    
		\includegraphics[width=1.0\columnwidth]{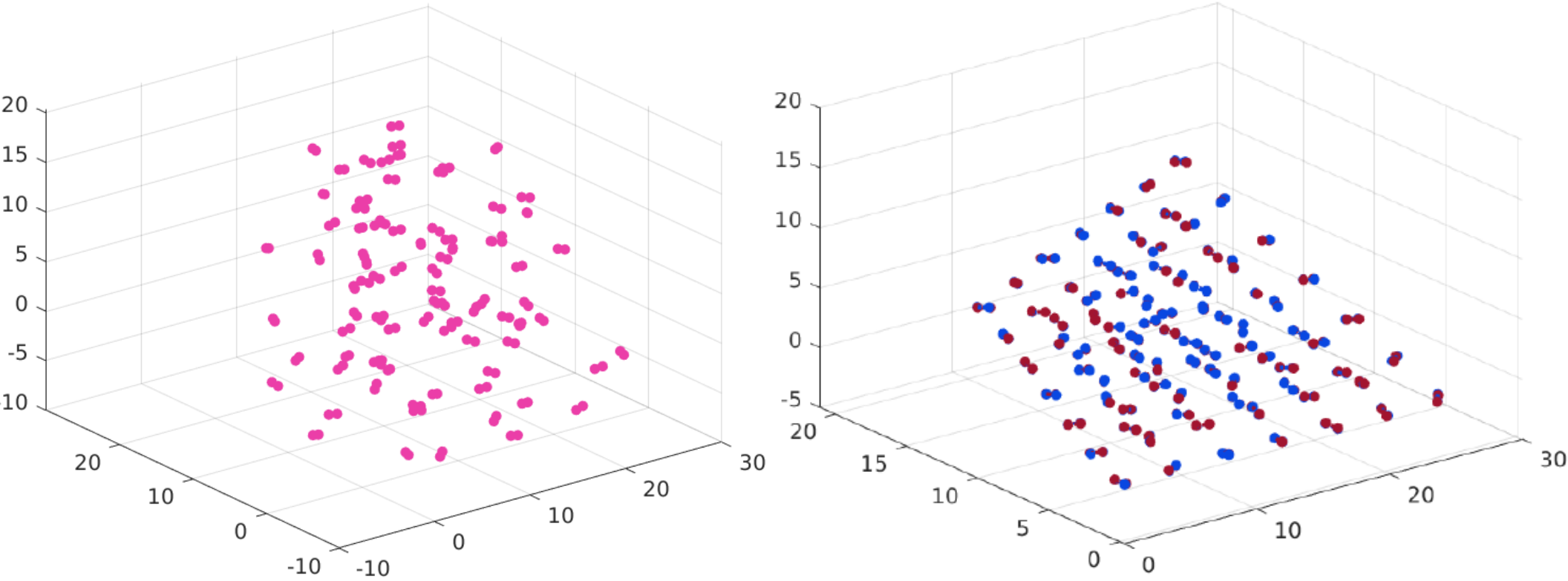}  
	} 
	\hspace{-0.25cm}\subfigure[Result before (left) and after (right) refinement of ESDP-BCD in \textbf{PYRAMID}] {    
		\includegraphics[width=1.0\columnwidth]{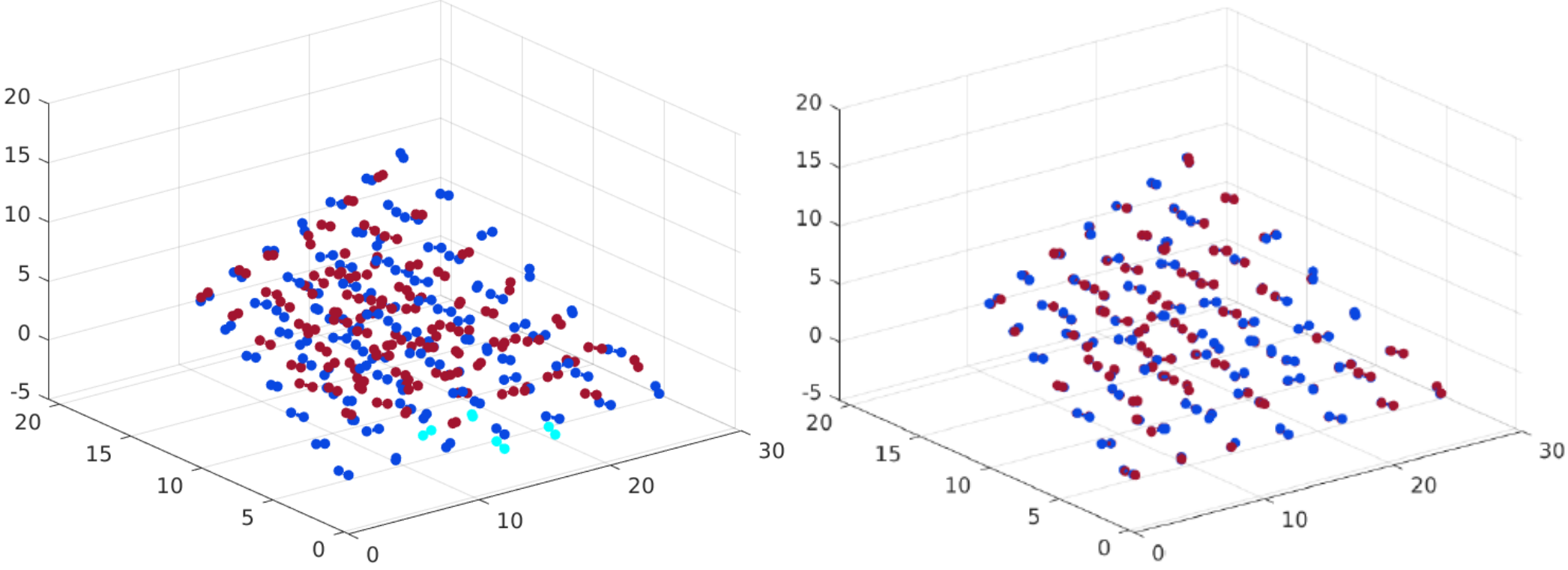}     
	}
	\hspace{-0.15cm}\subfigure[Initialization and result of BM-BCD($d$+1) in \textbf{RECTANGLE}] {    
		\includegraphics[width=1.0\columnwidth]{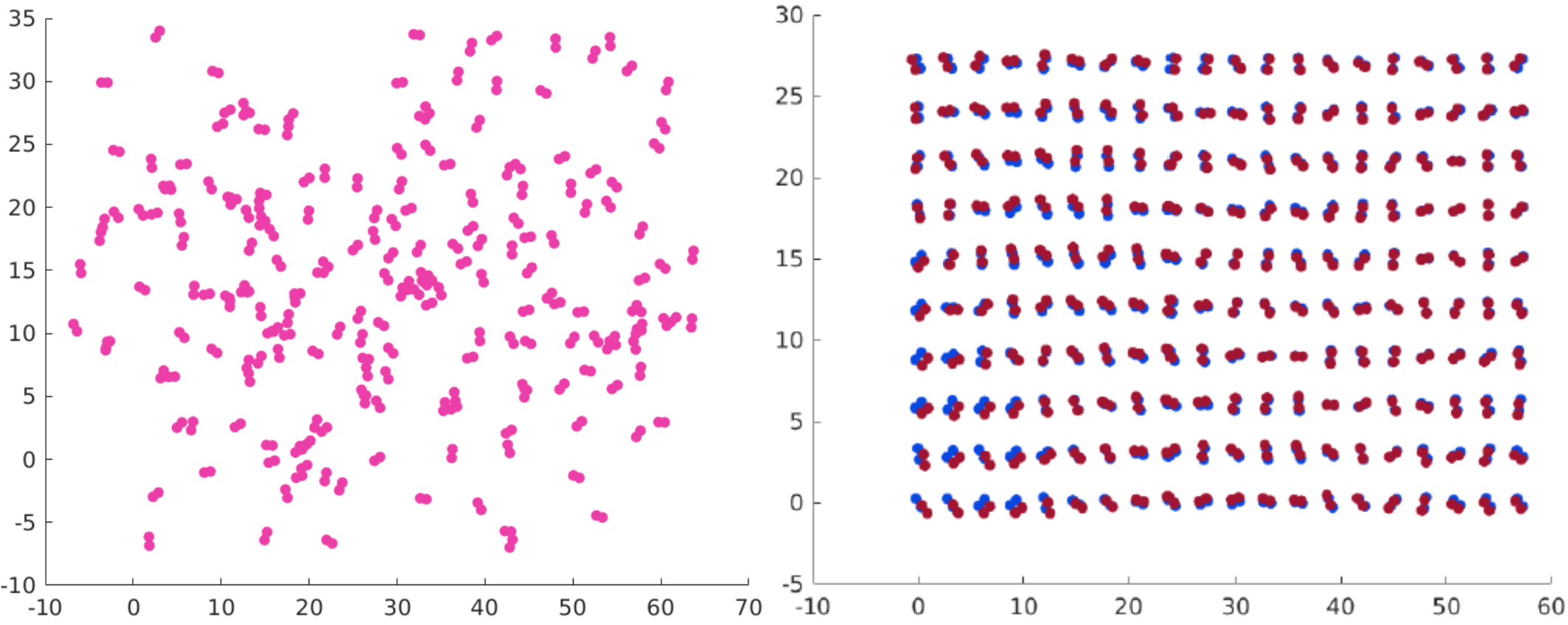}  
	} 
	\hspace{-0.25cm}\subfigure[Result before (left) and after (right) refinement of ESDP-BCD in \textbf{RECTANGLE}] {    
		\includegraphics[width=1.0\columnwidth]{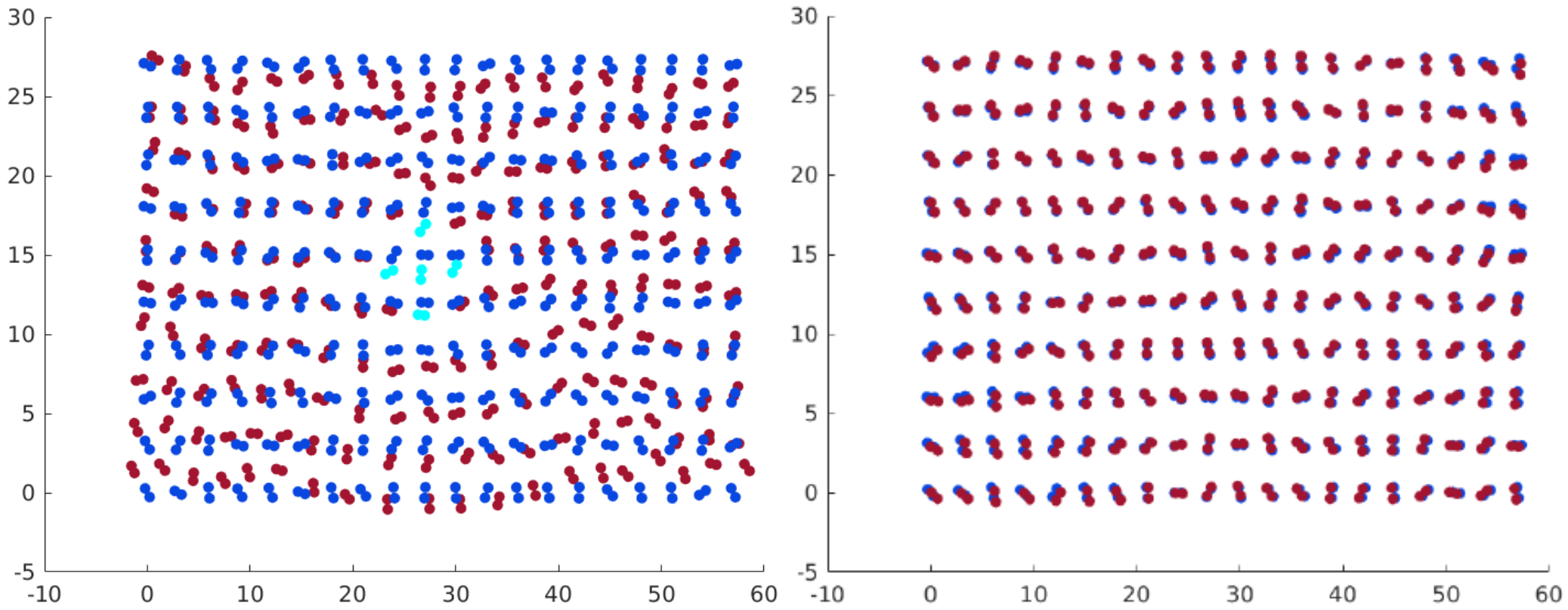}     
	}
	\footnotesize\textbf{Figure 2.} Visualization of the example problems with relative poor initializations in Table 2: the two connected balls together with the dotted line connecting them represent a robot, where the balls represent the two sensors on the robot.  \vspace{-0.4cm}      
\end{figure*}

\subsection{Choice of $r$ in BM-BCD}
We first empirically determine a reasonable parameter~$r$ according to the numerical results in Figure 1. In this experiment, we use poor initializations to trigger failures.

\textbf{Result.} It can be clearly seen from the results that lifting $r$ to a number larger than $d$ can make BM-BCD more robust than setting $r=d$ (i.e., refinement-only). As a cost of robustness, the number of iterations required by BM-BCD increases as $r$ rises, while the benefits in robustness to initializations from lifting $r$ seems to become limited. To tradeoff accuracy and efficiency, we believe that it is reasonable to take $d+1$ for BM-BCD.

\subsection{Noisy Measurements and Chaotic Initializations}
We evaluate the performance of our proposed methods in the four problems under \emph{poor} initializations. We show results for BM-BCD with setting $r=d+1$ as well as $r=d$ and ESDP-BCD with only 2$\sim$5 \emph{neighboring} anchors (see Figure 1), which is the toughest setup for ESDP. The default probability that an anchor has a measurement with one sensor of another robot is $0.3$.  The results are summarized in Table~2. We use the \mbox{$\left( \varDelta +1 \right)$-coloring} algorithm for block division to evaluate the total time (denoted as PT in Table 2) taken to solve the block update subproblem in a parallel and, for simplicity, synchronous manner for each BCD-type algorithms. We also show the total time (denoted as ST in Table 2) to solve each subproblem serially (i.e., in a column-wise block division). We use $k$ to denote the total iterations, where in ESDP-BCD we show iterations taken for "ESDP + refinement" repectively. To measure the accuracy of the relative state estimation, we use the average of the root-mean-squared-error (RMSE) for relative translations in \emph{body coordinate systems} for all tested methods. We define the failure rate (FR) as the rate at which RMSE is greater than 60 cm. 

\textbf{Result.} It can be seen from the results that 20$\sim$50 cm accuracy is commonly achievable in our setup with either BM-BCD or ESDP-BCD, of which the estimates are visualized in Figure 2. Among the NLP algorithms, BM-BCD($d+1$) combines  accuracy, robustness (as evidenced by FR), and efficiency, while performing slightly less accurately than QN at $d=2$ (probably due to the (quasi) second-order nature of the QN algorithm) but 1$\sim$2 orders of magnitude faster. Not surprisingly, both ESDP-BCD and SDP perform the best  robustness, both ESDP-BCD and SDP perform the best  robustness, since they can always provide an informed initialization for the refinement step. However, ESDP-BCD has a 2-order-of-magnitude reduction in computation time compared to SDP.

\begin{table}[]
	\label{table2}
	\setlength{\tabcolsep}{2pt}
	\renewcommand\arraystretch{1.1}
	\centering
	\hspace{-0cm}\begin{tabular}{cccc}
		\hline	Problem   & Method & \footnotesize{PT(s)} / \footnotesize{ST(s)} / $k$ / \footnotesize{$\rm RMSE$}(m) / \footnotesize{FR}  \\ \hline
		\footnotesize{\textbf{CUBE}} & BM-BCD($d$+1) & \textbf{0.12} / 0.43 / 34 / \textbf{0.27} / 0$\%$  \\
		$\rm \rho $ = 1/2, $\mathfrak{r}$ = 6m & BM-BCD($d$) & \textbf{0.08} / 0.30 / 19 / 0.30 / 0$\%$\\
		& QN & ------  / 96.5 / 98 / 0.32 / 0$\%$ \\
		& BCGD & 2.26 / 8.12 / 66 / $>$1 / 100$\%$ \\
		& ESDP-BCD & 2.88 / 12.9 / \textbf{3}+13 / \textbf{0.26} / 0$\%$ \\
		& SDP & ------ / $>$100 / ------ / \textbf{0.26} / 0$\%$ \\
		\footnotesize{\textbf{PYRAMID}} & BM-BCD($d$+1) & \textbf{0.11} / 0.35 / 39 / \textbf{0.38} / 2$\%$ \\
		$\rm \rho $ = 1/3, $\mathfrak{r}$ = 8m & BM-BCD($d$) & \textbf{0.08} / 0.25 / 24 / 0.55 / 16$\%$ \\
		& QN & ------ / 56.6 / 88 / 0.52 / 4$\%$ \\
		& BCGD & 2.03 / 6.11 / 80 / $>$1 / 100$\%$ \\
		& ESDP-BCD & 3.75 / 13.2 / \textbf{4}+16 / 0.43 / \textbf{0$\%$} \\
		& SDP & ------ / $>$100 / ------ / \textbf{0.37} / \textbf{0$\%$} \\
		\footnotesize{\textbf{HEXAGON}} & BM-BCD($d$+1) & \textbf{0.07} / 0.68 / 58 / 0.44 / 0$\%$ \\
		$\rm \rho $ = 1/4, $\mathfrak{r}$ = 8m & BM-BCD($d$) & \textbf{0.04} / 0.42 / 28 / 0.47 / 3$\%$  \\
		& QN & ------ / 88.2 / 83 / \textbf{0.42} / 0$\%$ \\
		& BCDG & 0.21 / 2.65 / 38 / 0.88 / 34$\%$ \\
		& ESDP-BCD & 1.88 / 17.9 / \textbf{3}+15 / \textbf{0.42} / 0$\%$ \\
		& SDP & ------ / $>$100 / ------ / \textbf{0.43} / 0$\%$ \\
		\footnotesize{\textbf{RECTANGLE}} & BM-BCD($d$+1) & \textbf{0.07} / 0.69 / 78 / 0.43 / 7$\%$ \\
		$\rm \rho $ = 1/3, $\mathfrak{r}$ = 9m & BM-BCD($d$) & \textbf{0.04} / 0.41 / 33 / 0.68 / 14$\%$ \\
		& QN & ------ / 75.3 / 82 / 0.42 / 2$\%$ \\
		& BCDG & unable to converge to desired precision   \\
		& ESDP-BCD & 0.73 / 11.3 / \textbf{3}+19 / \textbf{0.34} / \textbf{0$\%$}  \\
		& SDP & ------ / $>$100 / ------ / \textbf{0.34} / \textbf{0$\%$} \\ \hline
		\vspace{0.01mm}
	\end{tabular}
	
	\textbf{Table 2.} Numerical results of the proposed methods with alternatives. Statics are computed as the average results under 100 times of data generation for each problem.
\end{table}

\subsection{Initializations from informed to disordered}
\vspace{-0.1cm}
To visualize the complementary properties in terms of robustness and efficiency between the proposed methods, we illustrate the failure rate as well as \emph{serial} runtime (ST) when the initialization (quantified through $\rho$) varies. 

\textbf{Result.} From the Figure 3, it can be seen that BM-BCD($d+1$), despite outperforming other NLP algorithms in robustness and efficiency, inevitably yields to poor initializations. In contrast, the convex relaxation-based approach ESDP-BCD naturally avoids the sensitivity to initialization, at the expense of generally taking more time. We therefore believe that a pipeline as the cooperation of these two approaches, e.g., by emulating the front-end (BM-BCD) and back-end (ESDP-BCD) in SLAM systems, can provide practical computational solutions while achieving reasonable estimates.
\vspace{-0.2cm}
\begin{figure} \centering 
	\setcounter{subfigure}{0}   
	\includegraphics[width=1\columnwidth]{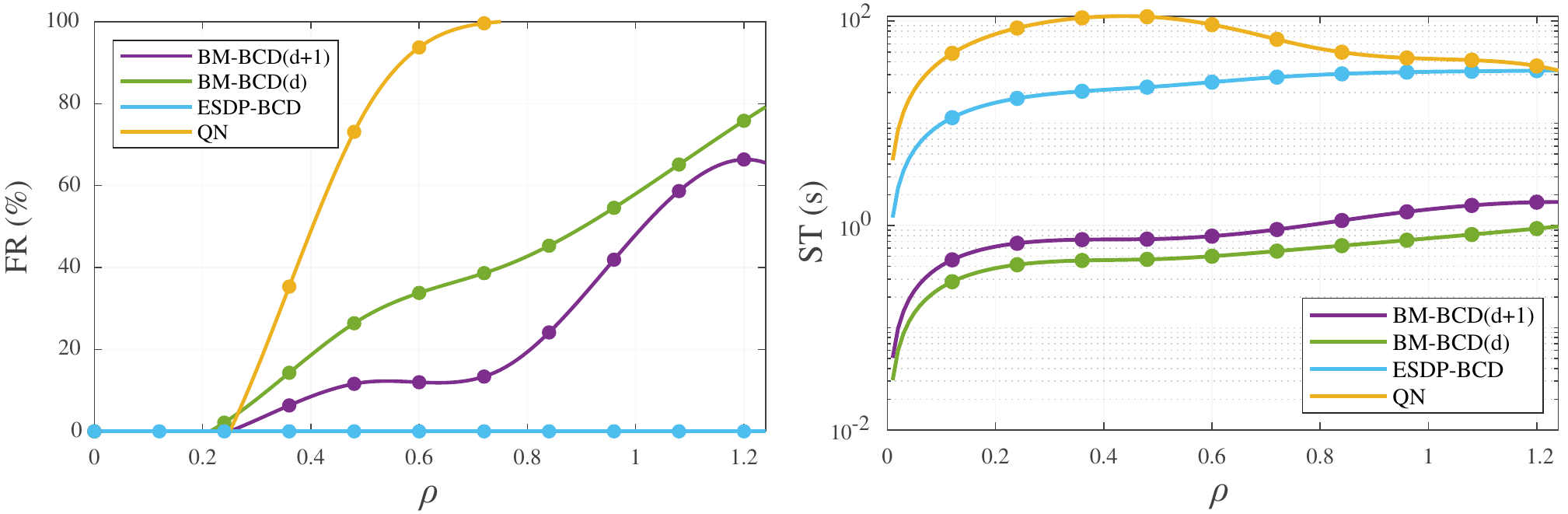}     
	\footnotesize\textbf{Figure 3.} FR and ST under varying initializations in \textbf{HEXAGON}. \vspace{-0.3cm}      
\end{figure} 

\begin{figure*} \centering
	\subfigure {  
		\includegraphics[width=0.58\columnwidth]{legend.pdf}  
	} \vspace{-0.3cm}\\   
	\setcounter{subfigure}{0} 
	\hspace{-0.15cm}\subfigure[Result before refinement of ESDP-BCD with 4 anchors available with \newline 0.1 (left) and 0.5 (right) anchor-measurement ratios] {
		\includegraphics[width=1.00\columnwidth]{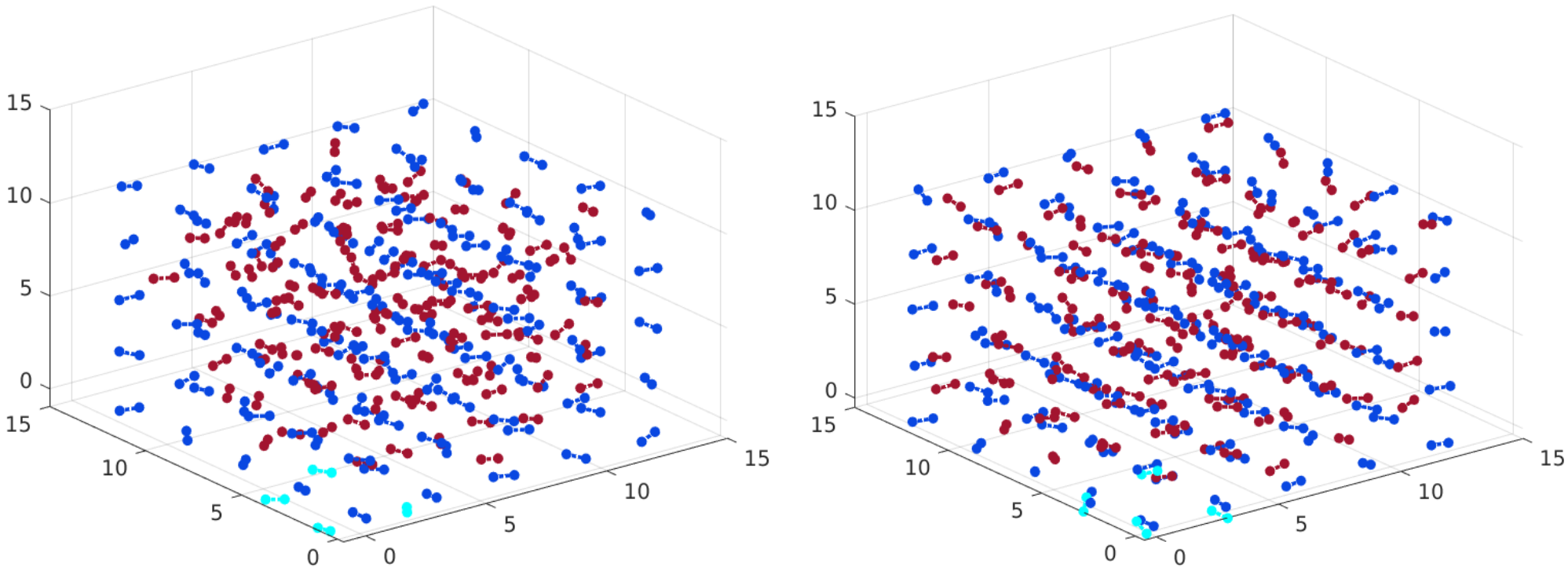}
	}      
	\hspace{-0.3cm}\subfigure[Result before refinement of ESDP-BCD with 8 anchors available with \newline 0.1 (left) and 0.5 (right) anchor-measurement ratios] {    
		\includegraphics[width=1.02    \columnwidth]{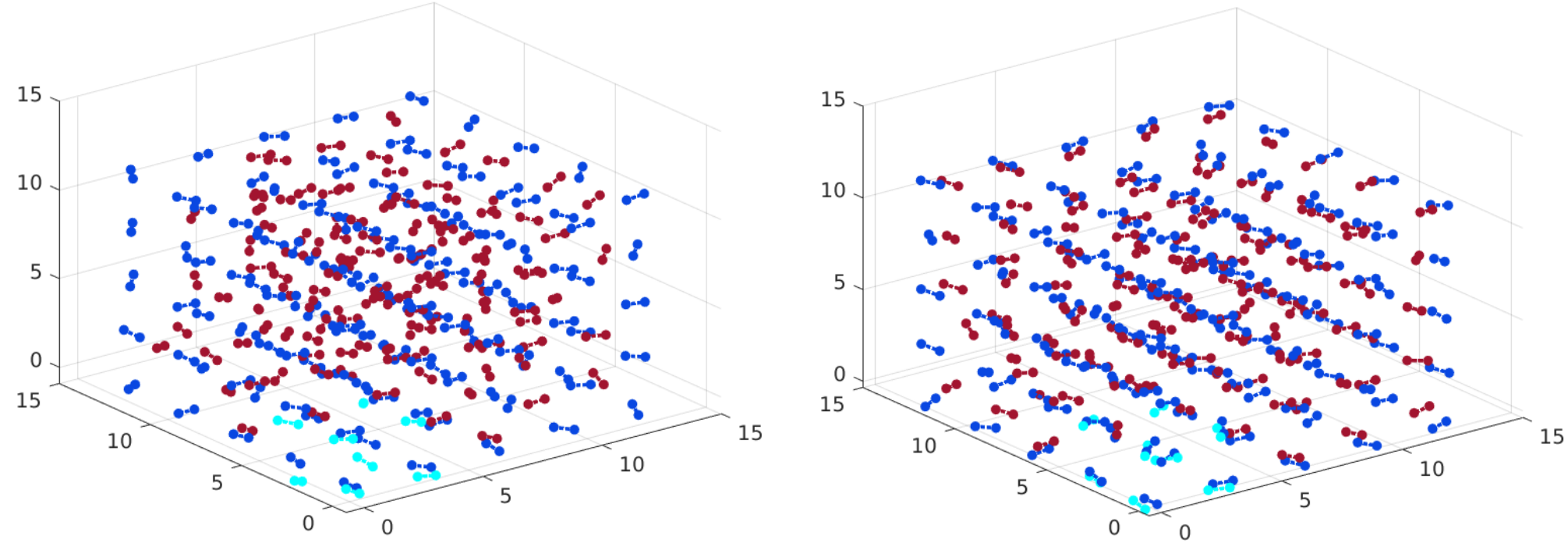}     
	} \\  
	\footnotesize\textbf{Figure 4.} Visualization of the results by ESDP-BCD with different anchor configurations in \textbf{CUBU}: the two connected balls together with the dotted line connecting them represent a robot, where the balls represent the two sensors on the robot.   
\end{figure*}

\begin{figure} \centering 
	\setcounter{subfigure}{0}   
	\includegraphics[width=1\columnwidth]{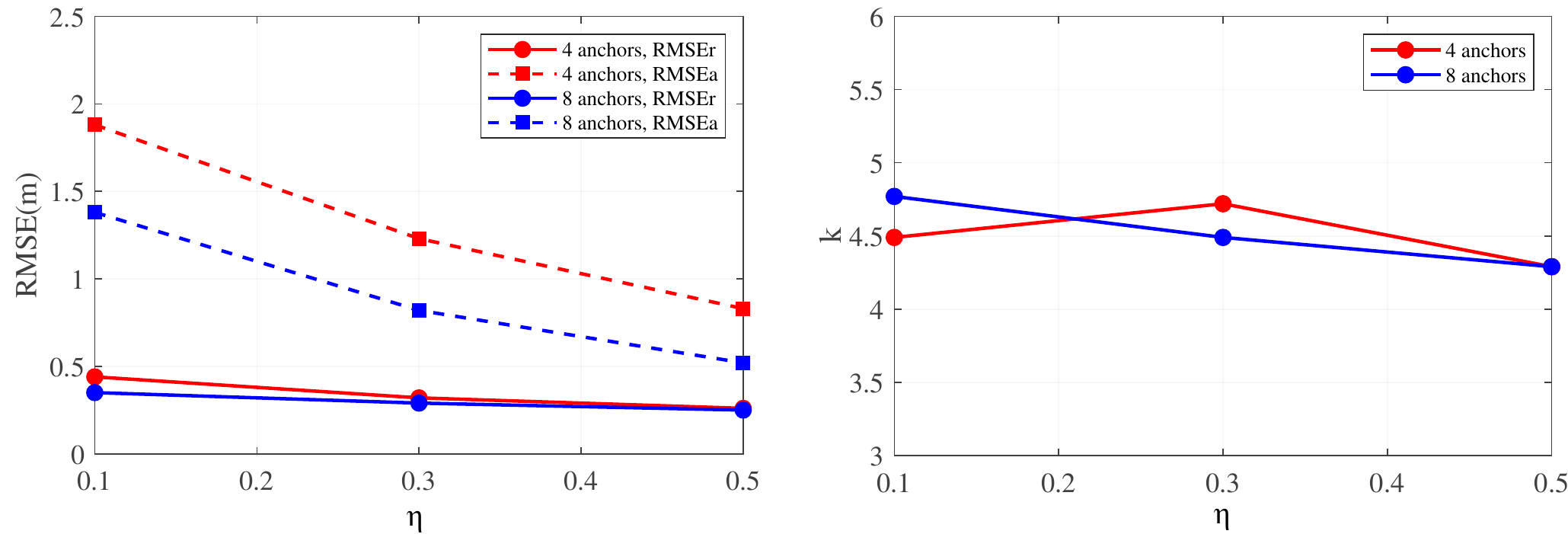}     
	\footnotesize\textbf{Figure 5.} RMSE and iterations $k$ under varying anchor-other measurement probability $\eta$ in \textbf{CUBE}. Statics are computed as the average results under 10 times of data generation. \vspace{-0.3cm}      
\end{figure}

\subsection{Varying configuration of anchors}
We convince the superiority of the proposed methods and intuitively revealed their complementary properties in the previous subsections. In this subsection, we explore the impact of numbers of anchors and  their measurement conditions, i.e., the configuration of anchors in ESDP-BCD. We show the results for different anchor configurations in the \textbf{CUBE} problem with $\rho=1$, from which a common patterns can be seen in other simulation problems.  We show in Figure~4 the visualization of the results for neighboring 4 anchors and 8 anchors. The corresponding accuracy of estimates and iterations ESDP-BCD takes are shown in Figure 5.  In Fig. 5 (left) we show the absolute estimation error of the robot's location before the refinement (note that unlike the RMSE shown previously, the error due to the robot's pose estimation is not taken into account here), denoted RMSE$_\text{a}$, and the relative error after the refinement, denoted RMSE$_\text{r}$, the same metrics as in the previous subsections.

\textbf{Result.} We can see from Figure 5 that 
when more (roughly localized) anchors are present, we only need a lower anchor-other measurement ratio $\eta$ to achieve the desired accuracy. Although the number of anchors in our experimental setup seems to affect little of the convergence rate (Fig. 5 (right)), without refinement, ESDP-BCD will produce more accurate results when more anchors are available. However, due to the robustness of BM-BCD to initializations, both configurations produce reasonable estimates after refinement, even if $\eta$ is set as 0.1. This provides us with practical guidance: if we can generate more anchors via small-scale relative state estimation, we should rely on fewer anchor-other measurements to achieve the desired accuracy, which alleviates the capacity/power requirements for individual distance sensors.

\section{Conclusion}
\label{sec:Conclusion}
\vspace{-0.1cm}
This paper proposes a complementary pair of distributed optimization scheme to handle the distance-based relative state estimation problem. The experimental results demonstrate that even in large-scale systems, our algorithms can still achieve real-time or near-real-time computational performance and result in reasonable estimates either with or without coarse initial guesses. This (at least at the algorithmic level) boosts our confidence in the use of distance measurements for scalable relative state estimation.\vspace{-0.1cm}

\newlength{\bibitemsep}\setlength{\bibitemsep}{0.00\baselineskip}
\newlength{\bibparskip}\setlength{\bibparskip}{0pt}
\let\oldthebibliography\thebibliography
\renewcommand\thebibliography[1]{
	\oldthebibliography{#1}
	\setlength{\parskip}{\bibitemsep}
	\setlength{\itemsep}{\bibparskip}
}
\bibliography{references}

\end{document}